\pdfoutput=1

\documentclass[11pt]{article}

\usepackage[]{acl}
\usepackage{times}
\usepackage{latexsym}
\usepackage{amsfonts,amsmath,amssymb,mathtools,bbm,bm}
\usepackage{amsthm}
\usepackage{graphicx}
\usepackage{subfig}
\theoremstyle{plain}

\newtheorem{proposition}{Proposition}

\newtheorem{assumption}{Assumption}

\usepackage[T1]{fontenc}

\usepackage[utf8]{inputenc}

\usepackage{microtype}

\usepackage{inconsolata}
\usepackage{multirow,multicol}
\usepackage{graphicx}
\usepackage{booktabs}
\usepackage{algpseudocode,algorithm,algorithmicx}
\usepackage{hyperref}
\title{Attention Alignment and Flexible Positional Embeddings \\ Improve Transformer Length Extrapolation}

\author{
  Ta-Chung Chi \\
  Carnegie Mellon University \\
  \texttt{tachungc@andrew.cmu.edu} \\\AND
  Ting-Han Fan \\
  Independent Researcher \\
  \texttt{tinghanf@alumni.princeton.edu} \\\And
  Alexander I. Rudnicky \\
  Carnegie Mellon University \\
  \texttt{air@cs.cmu.edu}
}

\begin{document}
\maketitle
\begin{abstract}
An ideal length-extrapolatable Transformer language model can handle sequences longer than the training length without any fine-tuning. 
Such long-context utilization capability relies heavily on a flexible positional embedding design. 
Upon investigating the flexibility of existing large pre-trained Transformer language models, we find that the T5 family deserves a closer look, as its positional embeddings capture rich and flexible attention patterns. 
However, T5 suffers from the dispersed attention issue: the longer the input sequence, the flatter the attention distribution. To alleviate the issue, we propose two attention alignment strategies via temperature scaling. 
Our findings show improvement on the long-context utilization capability of T5 on language modeling, retrieval, multi-document question answering, and code completion tasks without any fine-tuning. This suggests that a flexible positional embedding design and attention alignment can go a long way toward Transformer length extrapolation.\footnote{\url{https://github.com/chijames/Attention-Alignment-Transformer-Length-Extrapolation}}
\end{abstract}

\section{Introduction}
Pre-training large Transformer language models on long sequences is inherently expensive due to self-attention's quadratic complexity w.r.t the input sequence length~\cite{vaswani2017attention}. Even with the help of memory-efficient attention~\cite{rabe2021self, dao2022flashattention}, the maximum supported input length of current open-source pre-trained Transformer language models are capped at 4,096 tokens~\cite{touvron2023llama}, limiting their efficacy in handling long-context tasks.

\begin{table}[]
    \centering\resizebox{\linewidth}{!}{%
    \begin{tabular}{@{\extracolsep{3pt}}ccccccc}
    \hline\hline
    \multicolumn{7}{c}{\textbf{Retrieval Tasks}}\\
    \hline
    \multirow{2}{*}{Criteria} & \multicolumn{2}{c}{Topic} & \multicolumn{2}{c}{Line} & \multicolumn{2}{c}{Passkey} \\
    \cline{2-3} \cline{4-5} \cline{6-7}
     & 512 & 15k & 512 & 15k & 512 & 15k \\ \hline
    ${\rm P}_{\max}$ & 0.28 & 0.12 & 0.27 & 0.11 & 0.32 & 0.24 \\
    $\rm H$ & 3.47 & 6.63 & 3.47 & 7.04 & 3.09 & 5.97 \\
    \hline\hline
    \end{tabular}}
    \caption{\textbf{The Dispersed Attention Issue of Flan-T5-XL Encoder.} P$_{\max}$ is the average maximum probability and H is the average entropy. After increasing the sequence length from 512 to 15k, we observe larger entropy and smaller maximum probability, implying a flatter self-attention distribution.}
    \label{tab:ent_max_diff}
\end{table}

One notable research topic aiming to lift the input length restriction is~\emph{Length Extrapolation}~\cite{alibi}. 
Ideally, a length-extrapolatable Transformer language model is trained on short sequences and can perform equally well on longer ones without any further fine-tuning. This is made possible with carefully designed positional embeddings~\cite{alibi,chi2022kerple,sandwich}.
Unfortunately, existing approaches are tailored for natural language modeling, a task known to have strong recency bias, and they often do not perform well on other seemingly simple tasks such as passkey, topic, and line retrieval~\cite{mohtashami2023landmark,li2023long}.

\begin{table*}[ht!]
\centering\resizebox{\textwidth}{!}{%
\setlength{\tabcolsep}{2.5pt}
\begin{tabular}{@{}cccccccccc@{}}
\toprule
\textbf{Models} & T5~\shortcite{raffel2020exploring} & OPT~\shortcite{zhang2022opt} & ChatGLM~\shortcite{zeng2022glm} & LLaMA~\shortcite{touvron2023llama} & Falcon~\shortcite{penedo2023refinedweb} & Pythia~\shortcite{biderman2023pythia} & XGen~\shortcite{XGen} & BLOOM~\shortcite{scao2022bloom} & MPT~\shortcite{mpt} \\ \midrule
\multirow{2}{*}{\textbf{PE.}} & Learned & Learned & Rotary & Rotary & Rotary & Rotary & Rotary & ALiBi & ALiBi \\ 
& Relative & Absolute & Relative & Relative & Relative & Relative & Relative & Relative & Relative \\
\bottomrule
\end{tabular}
}
\caption{\textbf{Open-source Transformer language models and their positional embeddings.} T5 is the only model equipped with learnable relative positional embeddings, which enable its long-context utilization capability.}
\label{tab:models}
\end{table*}

To circumvent the recency bias, we sift through the positional embeddings of existing open-source large pre-trained Transformer language models, shown in Table~\ref{tab:models}, to find a flexible design, and the T5 family~\cite{raffel2020exploring} comes to our attention. As visualized in Figure~\ref{fig:t5_rpe_bias}, the flexibility of T5's positional embeddings allows it to encourage recency bias on one head and discourage that on another head. However, there is no free lunch: T5 suffers from the dispersed attention issue as shown in Table~\ref{tab:ent_max_diff}. That is, the attention distributions of long input sequences tend to be flatter than those of short input sequences. As a remedy, we propose two fine-tuning-free attention alignment strategies via Softmax temperature scaling~\cite{yao-etal-2021-self,kexuefm-8823} to mitigate the dispersed attention issue: maximum probability (${\rm P}_{\max}$) and entropy (${\rm H}$) alignment.

We validate the effectiveness of our alignment strategies on tasks including language modeling, retrieval, multi-document question answering, and code completion. 
We also provide a theoretical analysis of how the alignment strategies work under the hood by investigating the relation between the Softmax temperature and data distribution.

\section{Related Work}
\begin{figure*}[!htbp]
    \centering
    \subfloat[\centering 1$^{\text{st}}$ Attention Head]{{\includegraphics[width=0.4\linewidth]{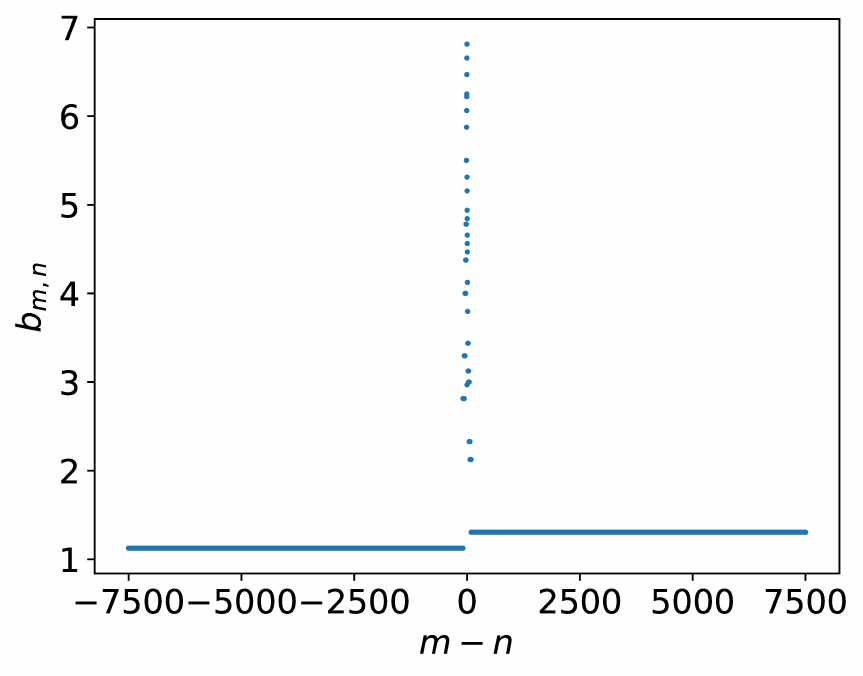}}}
    \quad\quad
    \subfloat[ \centering 27$^{\text{nd}}$ Attention Head]{{\includegraphics[width=0.425\linewidth]{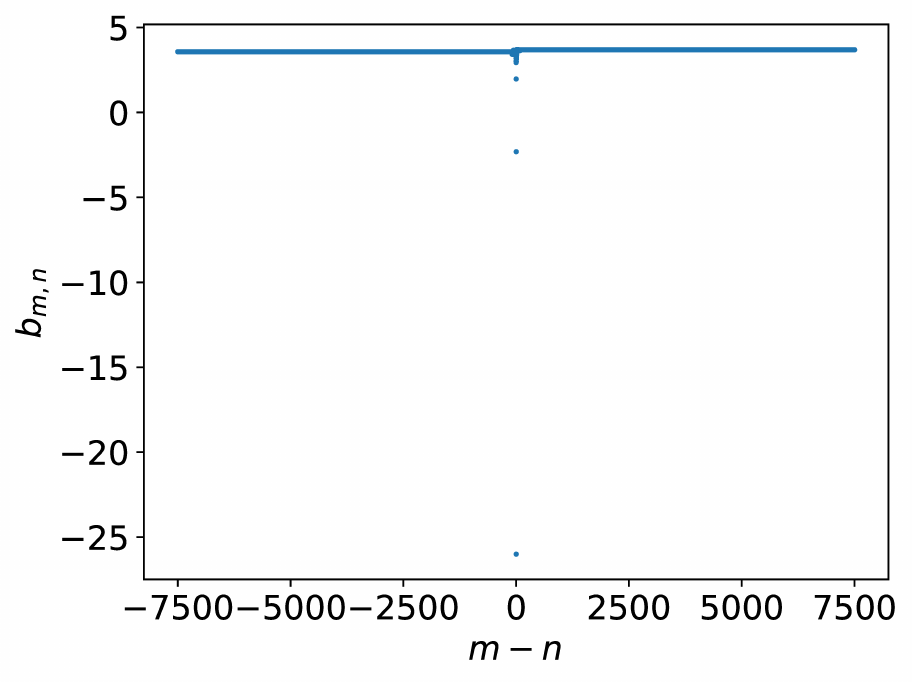}}}
    \caption{\textbf{Visualization of T5 Positional Embeddings.}  To plot figures of $b_{m,n}$, we set $m=7500$ and vary the value of $n$ from 0 to 15k. Each attention head of a Flan-T5-XL encoder learns a set of positional embeddings that capture different attention bias. For example, the positional embeddings in the left figure encourage the model to focus on nearby tokens. In contrast, the ones in the right figure let the model focus on only remote tokens.}
    \label{fig:t5_rpe_bias}
\end{figure*}

\paragraph{Transformer Positional Embeddings}
Transformer-based models rely on positional embeddings to encode positional information. We summarize open-source large pre-trained Transformer language models and their positional embeddings in Table~\ref{tab:models}. The relative variants are widely adopted due to their better empirical performance~\cite{rotary} and possible length-extrapolation capability~\cite{alibi}. In this work, we place special focus on the T5 positional embeddings due to their flexibility as shown in Figure~\ref{fig:t5_rpe_bias}.

\paragraph{Transformer Length Extrapolation}
Existing research on Transformer length extrapolation is mostly confined to the task of natural language modeling~\cite{alibi,chi2022kerple,sandwich}.
Unfortunately, the reported positive results do not carry over to long-context retrieval~\cite{mohtashami2023landmark,li2023long}. This contrastive observation can be explained by models' short empirical receptive field~\cite{sandwich}. In short, the strong decaying prior of positional embeddings prevents models from accessing distant tokens that may be necessary for retrieval tasks. In this work, we improve the flexible positional embeddings of T5 to get around this limitation.

\paragraph{Transformer Position Interpolation}
Instead of performing direct length extrapolation, a different line of research conducts model fine-tuning on long input sequences~\cite{chen2023extending}, where the main focus is to identify the most efficient fine-tuning scheme that can improve long-context utilization. Positive results have been reported on retrieval tasks~\cite{li2023long}. 
However, we argue that fine-tuning incurs additional costs since it needs 1) GPU resources to perform long sequence fine-tuning with large models and 2) a pre-defined target sequence length, which still imposes a sequence length upper limit. Our proposed methods can circumvent these two limitations.

\paragraph{Retrieval Tasks with Transformers}
Transformer-based approaches often consist of a retriever and a reader to overcome the context length restriction \cite{realm,rag,fid,retro}. The retriever retrieves relevant text snippets from a very large database and the reader digests the retrieved information to generate the correct output. Our proposed attention alignment strategy can be used to significantly increase the input sequence length of the reader, thereby allowing more retrieved information to participate in the decision process. For small-scale retrieval problems, our methods even obviate the need for context segmentation and the external key-value store used in prior work~\cite{mohtashami2023landmark}, serving as a more elegant approach.

\paragraph{Softmax Temperature Scaling}
To increase the length extrapolation capability of Transformers, previous work~\cite{yao-etal-2021-self,kexuefm-8823} scales the temperature of Softmax logarithmically w.r.t the sequence length to ensure invariant entropy. Our entropy alignment strategy is also inspired by this line of research except that we adopt a different procedure outlined below in Algorithm~\ref{alg:alignment}. Interestingly, our results in \S\ref{sec:temp_discussion} show that the logarithmic temperature scaling scheme is more similar to our proposed maximum probability alignment strategy.

\section{Long-context Retrieval Tasks with T5}
\label{sec:retrieval_t5}
\subsection{Why Retrieval?}
\label{sec:why_retrieval}
As suggested by recent work~\cite{mohtashami2023landmark,li2023long}, the task of long-context retrieval serves as a controllable benchmark to measure how well a Transformer language model utilizes long-context inputs.
One prominent characteristic of retrieval tasks is that only a subset of the input is of interest, requiring a model to accurately pick up the necessary information. The other characteristic is that the key information can sit anywhere in an input, requiring a model to attend flexibly. Finally, the controllable aspect allows us to gradually increase the input sequence length to test the models' length extrapolation capability.

\subsection{Why T5?}
To solve retrieval tasks using Transformer language models, it is necessary to choose a positional embedding design that permits accurate and flexible length-extrapolatable attention. After checking through the existing positional embeddings in Table~\ref{tab:models}, we find that the T5 family \cite{raffel2020exploring} fits our needs. 
As for other candidates, learnable absolute positional embeddings~\cite{vaswani2017attention,zhang2022opt} must be evaluated within the training length. ALiBi~\cite{alibi} and Rotary~\cite{rotary} have a recency bias;  they cannot extrapolate easily without fine-tuning.

For each attention head, T5 encoder maintains a bucket (B) of 32 learnable parameters and assigns the relative positional bias (rpe bias) $b_{m,n}$ as\footnote{\url{https://github.com/huggingface/transformers/blob/v4.33.2/src/transformers/models/t5/modeling\_t5.py\#L390}}
\begin{equation*}
\small
    \begin{split}
    b_{m,n}=&\\
&\hspace{-7mm}\begin{cases}
\text{B}[m-n],~\text{if~}0\leq m-n<8 \\
\text{B}[n-m+16],~\text{if~}-8<m-n<0 \\
\text{B}[\min(15,
8+\lfloor \frac{\log((m-n)/8)}{\log(128/8)}\cdot 8\rfloor)],~\text{if~}8\leq m-n\\
\text{B}[\min(31,
24+\lfloor \frac{\log((n-m)/8)}{\log(128/8)}\cdot 8\rfloor)],~\text{if~}m-n\leq -8,
\end{cases}
\end{split}
\end{equation*}
where $0\leq m<L$ and $0\leq n<L$ are two position indices. $b_{m,n}$ will be added to the $(m,n)$-th entry of the $L\times L$ self-attention matrix. The summation becomes the input to the temperature-scaled Softmax.
We plot the learned rpe bias of a T5 encoder in Figure~\ref{fig:t5_rpe_bias}. We can tell that its attention heads encode rich attention patterns. For example, head 1 learns to focus on the nearby tokens whereas head 27 learns to ignore the nearby tokens and allow access to faraway tokens.

\subsection{The Dispersed Attention Issue of T5 Encoder}
Unfortunately, directly applying T5 models on retrieval tasks does not yield perfect results. Upon inspecting the intermediate model states, we find that a longer input sequence consists of more tokens competing for the same amount (i.e., Softmax sums to 1) of attention, resulting in the dispersed attention issue.
In Table~\ref{tab:ent_max_diff}, we see that the longer the input sequence, the flatter the self-attention distribution. The situation is not hopeless if the desired information still attains a higher attention weight than the remaining tokens. Our proposed solution in \S\ref{sec:proposed_methods} will let the key information stand out.

\begin{algorithm}
\small
\caption{Attention Alignment Strategies}\label{alg:alignment}
\begin{algorithmic}
\Require A short sequence of length $L_{tr}$ and a long sequence of length $L_{ex} > L_{tr}$. Encoder $E$. Alignment mode $M$.
\Ensure The Softmax temperature $\tau$
\Function{FindS}{$\tau$, $M$}
\State Set temperature of all Softmax to $\tau$
\State $s \gets [~]$
\For{operation in $E$}
\State Perform the operation
\If{operation is $\text{Softmax}_{\tau}(l)$}
\If{$M$ is Maximum Probability}
    \State Append $\max(\text{Softmax}_{\tau}(l))$ to $s$
\ElsIf{$M$ is Entropy}
\State Append ${\rm H}(\text{Softmax}_{\tau}(l))$ to $s$
\EndIf
\EndIf
\EndFor
\State \Return $\text{avg}(s)$
\EndFunction

\State $\overline{S^{tr}}(1)\gets \textproc{FindS}(1.0, M)$

\For{$\tau_{ex}=1.0, 0.95, 0.9,\cdots,0.5$}
\State $\overline{S^{ex}}(\tau_{ex})=\textproc{FindS}(\tau_{ex}, M)$
\EndFor
\State \Return $\tau_{ex}$ s.t. $\overline{S^{ex}}(\tau_{ex}) \approx \overline{S^{tr}}(1)$
\end{algorithmic}
\end{algorithm}

\section{Proposed Methods}
\label{sec:proposed_methods}
A natural solution to the dispersed attention issue described in \S\ref{sec:retrieval_t5} is to sharpen the self-attention distribution. This can be achieved by reducing the temperature $\tau$ during extrapolation. We set the extrapolation temperature $\tau_{ex}$ such that the sharpness during training with $\tau_{tr}=1$ and that during extrapolation with $\tau_{ex}<1$ are roughly the same. As a measurement of sharpness, we explore the maximum probability or entropy of a distribution. In other words, our proposed solution is to align the maximum probability or entropy of training and extrapolation distributions by adjusting $\tau_{ex}$.

Concretely, let $l^{(i)}\in\mathbb{R}^L$ be the i-th pre-Softmax logit vector of a T5 encoder, where $L\in\{L_{tr}, L_{ex}\}$ is the sequence length. The post-Softmax distribution of $l^{(i)}$ is ${\rm P}^{(i)}(\tau)=\text{Softmax}_{\tau}(l^{(i)})$. The maximum probability and entropy of ${\rm P}^{(i)}(\tau)$ are ${\rm P}_{\max}^{(i)}(\tau)$ and ${\rm H}^{(i)}(\tau)$, respectively.

Take the maximum probability alignment strategy as an example: We first run the forward pass and compute the average maximum probability under temperature $\tau$ over all logit vectors: $\overline{{\rm P}_{\max}}(\tau)=(1/N)\sum_i {\rm P}_{\max}^{(i)}(\tau)$ where $N=R\times H\times L$ is the number of logit vectors in a T5 encoder with $R$ layers, $H$ heads, and length-$L$ sequences. Since the temperature is 1 during training and $\tau_{ex}$ during extrapolation, we denote the average maximum probability during training as $\overline{{\rm P}_{\max}^{tr}}(1)$ and that during extrapolation as $\overline{{\rm P}_{\max}^{ex}}(\tau_{ex})$. Finally, to align the maximum probabilities, we adjust $\tau_{ex}$ s.t. $\overline{{\rm P}_{\max}^{ex}}(\tau_{ex}) \approx \overline{{\rm P}_{\max}^{tr}}(1)$. In practice, we do a grid search on $\tau_{ex}$ from 1.0 to 0.5. We outline the procedure of the alignment strategies in Algorithm~\ref{alg:alignment}.

Note that our proposed methods do not require any model fine-tuning or gradient computations. The only overhead is estimating the temperature $\tau_{ex}$ using Algorithm~\ref{alg:alignment} and a few length $L_{ex}$ sequences. Once the temperature is decided, it will be held fixed, rendering our methods simple and efficient. In addition, our fine-tuning free methods do not lead to performance regression on short $L_{tr}$ sequences commonly observed on long-context fine-tuned models~\cite{roziere2023code}.

\section{Experiments}
\label{sec:experiments}
We compare the two alignment strategies against the length-only Softmax temperature scaling scheme $\tau=
\log_{L_{ex}}L_{tr}$~\cite{yao-etal-2021-self,kexuefm-8823} and LongChat-13B-16K~\cite{li2023long}. Note that LongChat-13B-16K~\cite{li2023long}, the best baseline, was fine-tuned from LLaMA~\cite{touvron2023llama} on long sequences of length 16k while our proposed methods do not need any fine-tuning.

\begin{table}[!ht]
    \setlength{\tabcolsep}{1.2pt}
    \centering
    \scalebox{0.9}{
    \begin{tabular}{@{\extracolsep{1pt}}lcccccc}
    \hline\hline
    \multicolumn{7}{c}{\textbf{Language Modeling}}\\
    \hline
    \multirow{2}{*}{Models} & \multicolumn{5}{c}{Sequence Length ($L_{ex}$)} \\
    \cline{2-7}
     & 1024 & 2048 & 4096 & 8192 & 15000 & Avg. \\ \hline 
    T5-Large-LM & 35.9 & 40.1 & >1k & >1k & >1k  & > 1k\\
    w/ ${\rm P}_{\max}$ & 34.7 & 45.5 & 45.2 & 45.5 & 52.7 & \textbf{44.7} \\
    w/ $\rm H$ & 40.2 & 43.9 & 45.6 & 54.6 & 56.0 & 48.1 \\
    w/ $\log_{L_{ex}} L_{tr}$ & 39.8 & 38.2 & 47.4 & 45.3 & 55.9 & 45.3 \\ \hline
    T5-XL-LM & 28.3 & >1k & >1k & >1k & >1k & > 1k\\
    w/ ${\rm P}_{\max}$ & 30.2 & 36.0 & 31.6 & 41.7 & 50.0 & 37.9 \\
    w/ $\rm H$ & 30.4 & 36.8 & 38.4 & 53.3 & 63.4 & 44.4 \\
    w/ $\log_{L_{ex}} L_{tr}$ & 27.3 & 29.4 & 31.7 & 39.3 & 45.8 & \textbf{34.7} \\ \hline
    T5-XXL-LM & 109 & >1k & >1k & >1k & >1k & > 1k \\
    w/ ${\rm P}_{\max}$ & 32.2 & 29.7 & 29.5 & 36.6 & 44.3 & 34.5 \\
    w/ $\rm H$ & 26.8 & 28.1 & 34.2 & 37.8 & 43.8 & \textbf{34.1} \\
    w/ $\log_{L_{ex}} L_{tr}$ & 27.1 & 36.1 & 33.9 & 246 & 43.8 & 77.5 \\
    \hline\hline
    \end{tabular}}
    \caption[abc]{\textbf{Language Modeling Performance.} We report the average perplexity of 500 sequences. The lower the better.}
    \label{tab:lm_exp}
\end{table}

\begin{table*}[!ht]
    \setlength{\tabcolsep}{2.3pt}
    \centering
    \scalebox{0.9}{
    \begin{tabular}{@{\extracolsep{1.5pt}}lccccccccccccccccc}
    \hline\hline
    \multicolumn{18}{c}{\textbf{Retrieval Tasks}}\\
    \hline
    \multirow{2}{*}{Models} & \multicolumn{5}{c}{Topic, \# of topics} & \multicolumn{6}{c}{Line, \# of lines} & \multicolumn{5}{c}{Passkey, \# of sentences} & \multirow{2}{*}{Avg.} \\
    \cline{2-6}\cline{7-12}\cline{13-17}
     & 5 & 10 & 15 & 20 & 25 & 200 & 300 & 400 & 500 & 600 & 680 & 20k & 30k & 40k & 50k & 55k \\ \hline
    Flan-T5-Large & 99 & 100 & 97 & 97 & 83 & 97 & 100 & 92 & 96 & 93 & 92 & 62 & 47 & 31 & 16 & 9 & 76 \\
    w/ ${\rm P}_{\max}$ & 96 & 90 & 86 & 94 & 98 & 99 & 98 & 98 & 98 & 98 & 100 & 84 & 90 & 85  & 79 & 85 & \textbf{92} \\
    w/ $\rm H$ & 59 & 32 & 16 & 0 & 3 & 97 & 90 & 94 & 83 & 93 & 88 & 29 & 25 & 21 & 15 & 22 & 48 \\
    w/ $\log_{L_{ex}} L_{tr}$ & 88 & 79 & 75 & 61 & 55 & 99 & 99 & 98 & 99 & 97 & 98 & 74 & 63 & 51 & 41 & 35 & 76 \\ \hline
    Flan-T5-XL & 100 & 100 & 100 & 100 & 100 & 96 & 90 & 77 & 57 & 45 & 26 & 100 & 100 & 100 & 100 & 100 & 87 \\
    w/ ${\rm P}_{\max}$ & 100 & 100 & 100 & 100 & 100 & 97 & 90 & 89 & 80 & 70 & 62 & 100 & 99 & 100 & 100 & 100 & \textbf{93} \\
    w/ $\rm H$ & 99 & 98 & 97 & 96 & 96 & 95 & 87 & 88 & 79	& 70 & 71 & 100 & 100 & 100 & 100 & 100 & 92 \\
    w/ $\log_{L_{ex}} L_{tr}$ & 99 & 100 & 100 & 100 & 100 & 98 & 88 & 81 & 86 & 60 & 67 & 100 & 100 & 100 & 100 & 99 & 92 \\ \hline
    Flan-T5-XXL & 100 & 100 & 100 & 99 & 99 & 100 & 100 & 98 & 95 & 84 & 82 & 100 & 100 & 100 & 100 & 100 & 97 \\
    w/ ${\rm P}_{\max}$ & 100 & 100 & 100 & 99 & 99 & 97 & 99 & 96 & 97 & 94 & 95 & 100 & 98 & 100 & 100 & 100 & \textbf{98} \\
    w/ $\rm H$ & 100 & 100 & 97 & 98 & 94 & 99 & 92 & 92 & 76 & 58 & 58 & 100 & 100 & 100 & 100 & 100 & 92 \\
    w/ $\log_{L_{ex}} L_{tr}$ & 100 & 100 & 99 & 98 & 92 & 100 & 98 & 94 & 93 & 84 & 90 & 100 & 100 & 100 & 100 & 100 & 97 \\ \hline
    LongChat & 100 & 100 & 100 & 99 & 89 & 100 & 91 & 93 & 83 & 78 & 59 & 100 & 100 & 99 & 100 & 99 & 93 \\
    \hline\hline
    \end{tabular}}
    \caption{\textbf{Performance of Retrieval Tasks.} The numbers are accuracy. Full score is 100. The LongChat model corresponds to the LongChat-13B-16K model~\cite{li2023long}. It is a LLaMA-13B~\cite{touvron2023llama} model fine-tuned on sequences of length 16k using the positional interpolation technique~\cite{chen2023extending}. Flan-T5-XXL has 11B parameters. The maximum sequence lengths ($L_{ex}$) of the three tasks are all around 14.5k to 15.5k tokens.}
    \label{tab:retrieval_exp}
\end{table*}

\subsection{Language Modeling}
We use the LM-Adapted T5 models for this experiment\footnote{\url{https://github.com/google-research/text-to-text-transfer-transformer/blob/main/released_checkpoints.md\#lm-adapted-t511lm100k}}. We set $L_{tr}=512$.
Following previous work on Transformer length extrapolation, we perform an intrinsic evaluation on language modeling~\cite{alibi,chi2022kerple,sandwich}.
Ideally, our proposed methods should alleviate the perplexity explosion problem during extrapolation. As we can see in Table~\ref{tab:lm_exp}, both alignment strategies dramatically improve (lower) the perplexity. We also observe that scaling the temperature solely based on sequence lengths is not the optimal strategy, as indicated by the sudden perplexity increase of the $\log_{L_{ex}}L_{tr}$ strategy. We will provide an in-depth discussion on this topic in \S\ref{sec:temp_discussion}.
Note that perplexity is not our primary focus since it often cannot accurately reflect the long-context utilization capability of Transformers on practical tasks~\cite{li2023long}.

\subsection{Long-context Retrieval}
\label{sec:retrieval_tasks}
The tasks are formulated in the Question Answering (QA) format; therefore, we use the Flan-T5 models to leverage their instruction-following capability. We set $L_{tr}=512$.
Inspired by recently proposed retrieval tasks, we evaluate the proposed alignment strategies on three of these. Topic retrieval requires a model to retrieve the first topic in a long and multi-topic conversation~\cite{li2023long}. Line retrieval has a long series of key-value pairs, and a model needs to retrieve the value corresponding to the given key~\cite{li2023long}. Passkey retrieval hides a passkey in a long junk text snippet, and a model needs to return that passkey~\cite{mohtashami2023landmark}.

As we can see in Table~\ref{tab:retrieval_exp}, the retrieval performance is greatly boosted after the Flan-T5 models are equipped with our proposed attention alignment strategies. In particular, the maximum probability alignment strategy provides better results across the board. Other baselines such as MPT~\cite{mpt} and ChatGLM2~\cite{du2022glm} perform worse than LongChat. Please refer to~\citet{li2023long} for more details.
We also present the optimal temperature given by Algorithm~\ref{alg:alignment} in Table~\ref{tab:retrieval_temps} in Appendix~\ref{sec:appendix_temps}. In short, the temperature decreases when the input sequence length increases. We will provide additional temperature analysis below, in \S{\ref{sec:temp_discussion}.

\begin{table*}[!ht]
    \setlength{\tabcolsep}{3pt}
    \centering
    \scalebox{0.9}{
    \begin{tabular}{@{\extracolsep{6pt}}lcccccccccc}
    \hline\hline
    \multicolumn{11}{c}{\textbf{Multi-document Question Answering}}\\
    \hline
    \multirow{2}{*}{Models} & \multicolumn{1}{c}{10 Docs} & \multicolumn{1}{c}{20 Docs} & \multicolumn{8}{c}{30 Docs, golden doc at different positions} \\
    \cline{2-2} \cline{3-3} \cline{4-11}
     & Avg. & Avg. & 0 & 4 & 9 & 14 & 19 & 24 & 29 & Avg. \\ \hline
    Flan-T5-Large & 52.4 & 43.3 & 52.6 & 42.0 & 36.5 & 34.0 & 33.9 & 33.9 & 37.9 & 38.7 \\
    w/ ${\rm P}_{\max}$ & \textbf{53.1} & \textbf{44.2} & 50.8 & 44.5 & 39.5 & 36.4 & 35.9 & 35.8 & 37.0 & \textbf{40.0} \\
    Improvement & +0.7 & +0.9 & -1.8 & +1.5 & +3.0 & +2.4 & +2.0 & +1.9 & -0.9 & +1.3 \\
    w/ $\rm H$ & 52.1 & 43.2 & 47.6 & 41.1 & 35.2 & 33.5 & 32.2 & 33.3 & 34.2 & 36.7 \\ 
    w/ $\log_{L_{ex}} L_{tr}$ & 53.2 & 44.5 & 50.6 & 44.1 & 39.3 & 36.3 & 35.8 & 35.8 & 37.2 & 39.9 \\
    \hline
    Flan-T5-XL & 59.4 & 51.2 & 58.4 & 44.6 & 40.0 & 39.9 & 41.7 & 46.4 & 54.8 & 46.5 \\
    w/ ${\rm P}_{\max}$ & \textbf{61.1} & \textbf{53.6} & 60.9 & 49.1 & 46.0 & 44.9 & 46.3 & 49.1 & 55.7 & \textbf{50.3} \\
    Improvement & +1.7 & +2.4 & +2.5 & +4.5 & +6.0 & +5.0 & +4.6 & +2.7 & +0.9 & +3.8 \\
    w/ $\rm H$ & 60.5 & 52.4 & 52.4 & 43.5 & 42.1 & 40.3 & 42.0 & 42.9 & 51.3 & 44.9 \\
    w/ $\log_{L_{ex}} L_{tr}$ & 60.9 & \textbf{53.6} & 61.0 & 49.1 & 46.1 & 44.7 & 46.1 & 48.7 & 55.4 & 50.2 \\
    \hline
    Flan-T5-XXL & 63.6 & 56.9 & 58.9 & 49.1 & 48.1 & 47.5 & 48.9 & 53.1 & 61.2 & 52.4 \\
    w/ ${\rm P}_{\max}$ & 63.7 & \textbf{57.7} & 60.4 & 52.5 & 51.0 & 50.2 & 51.3 & 53.5 & 59.1 & \textbf{54.0} \\
    Improvement & +0.1 & +0.8 & +1.5 & +3.4 & +2.9 & +2.7 & +2.4 & +0.4 & -2.1 & +1.6 \\
    w/ $\rm H$ & 63.6 & 57.1 & 61.0 & 53.4 & 50.8 & 50.3 & 50.7 & 51.9 & 55.7 & 53.4 \\
    w/ $\log_{L_{ex}} L_{tr}$ & \textbf{63.9} & 57.6 & 61.5 & 53.3 & 51.3 & 50.3 & 51.1 & 53.0 & 57.2 & \textbf{54.0} \\
    \hline\hline
    \end{tabular}}
    \caption{\textbf{Performance of Multi-document QA.} Numbers are accuracy. Full score is 100. The maximum sequence length ($L_{ex}$) of 30 documents is around 5k. The improvement row represents the absolute accuracy improvement after a Flan-T5 model is equipped with our proposed maximum probability alignment strategy. For the full performance breakdown, please refer to Table~\ref{tab:qa_exp_full} in Appendix~\ref{sec:appendix_qa_exp_full}.}
    \label{tab:qa_exp}
\end{table*}

\subsection{Multi-document Question Answering}
We again use the Flan-T5 models to leverage their instruction-following capability. We set $L_{tr}=512$.
We choose the multi-document question-answering task as our downstream task~\cite{liu2023lost}.
The model input consists of a question Q and multiple documents extracted from NaturalQuestions~\cite{kwiatkowski2019natural} related to Q, where one of the documents (golden doc) contains the ground truth answer to Q. As shown in Table~\ref{tab:qa_exp}, when a model is equipped with the proposed maximum probability alignment strategy, it consistently outperforms the original model across model sizes and number of input documents.

Apart from the better task performance, we believe that the attention dispersed attention issue discussed in \S\ref{sec:retrieval_t5} can help demystify the lost-in-the-middle phenomenon~\cite{liu2023lost} of this task: Transformer models tend to perform worse when the ground truth sits near the middle of the input context. Let us recall the relative positional embedding of head 27 learned in Figure~\ref{fig:t5_rpe_bias}, if the ground truth answer sits in the middle, it will have long contexts from both sides competing for the attention weight. If this hypothesis is correct, we can expect the performance boost to be more prominent when the answer appears near the middle. We reveal the performance breakdown when the number of input documents is 30. As we can see in the improvement row, those cases indeed achieve greater improvements.

Our strategies are not always perfect: The performance drops if the ground truth answer is at position 29. We believe T5 might have already handled this case pretty well due to the recency bias learned on some attention heads, and our additional temperature scaling sharpens the distribution too aggressively. We acknowledge this trade-off in \S\ref{sec:limitation}.

\subsection{Code Key Retrieval and Completion}
To test the generalizability of the alignment strategies, we apply our methods to the CodeT5+ model~\cite{wang2023codet5+} that was pre-trained on code data with 770M parameters.\footnote{\url{https://huggingface.co/Salesforce/codet5p-770m-py}} We set $L_{tr}=768$. We do not experiment with larger CodeT5+ models since they do not follow the T5 architecture, but use other positional embeddings. We conduct two experiments on the LCC dataset~\cite{Guo2023LongCoderAL}, which is highly similar to the classic PY150 dataset~\cite{raychev2016probabilistic} except that the input context length is much longer. 

For the code key retrieval experiment, we sample several code files from LCC along with a special function that only returns an integer from 1 to 100. We concatenate them and ask a model to generate the returned integer at the end~\cite{roziere2023code}. Considering that this is essentially a passkey retrieval task in the code domain, we briefly report the average accuracy of 100 test cases when the input sequence length is around 16k: 0 (Original CodeT5+), \textbf{87} (w/ ${\rm P}_{\max}$), 80 (w/ $\rm H$), and 85 (w/ $\log_{L_{ex}} L_{tr}$). We can see that the maximum probability alignment strategy performs the best.

For the code completion experiment, a model needs to generate the next line of code given some prior code as the context. The metrics are Exact Match (EM) and Edit similarity (ES) on a per line basis~\cite{svyatkovskiy2020intellicode}. We report the results in Table~\ref{tab:code_exp_truncated} using the context length bucketing format. While both alignment strategies improve the performance substantially, ${\rm P}_{\max}$ is better; however, its EM performance lags behind $\log_{L_{ex}} L_{tr}$ when the sequence length increases.
We additionally include an extrapolation-free baseline, \emph{truncation}, that truncates the long input context to the most recent $L_{tr}=768$ tokens. Both ${\rm P}_{\max}$ and $\log_{L_{ex}} L_{tr}$ perform better than this baseline when $L_{ex} < 6000$, indicating that they can indeed benefit from longer ($6000/768=7.8$x) contexts without any fine-tuning.

\begin{table*}[!ht]
    \setlength{\tabcolsep}{2pt}
    \centering
    \scalebox{0.9}{
    \begin{tabular}{@{\extracolsep{2pt}}lcccccc}
    \hline\hline
    \multicolumn{7}{c}{\textbf{Code Completion Exact Match}}\\
    \hline
    \multirow{2}{*}{Models} & \multicolumn{6}{c}{Sequence Length ($L_{ex}$)} \\
    \cline{2-7}
     & 1k & 2k & 3k & 4k & 5k & 6k \\ \hline
    CodeT5+ & 19.6 & 19.0 & 11.3 & 2.6 & 0.1 & 0.0 \\
    w/ ${\rm P}_{\max}$ & 21.1 & 22.5 & 21.7 & 21.5 & 19.3 & 22.7 \\
    w/ $\rm H$ & 19.5 & 18.7 & 13.7 & 9.0 & 7.9 & 9.0 \\
    w/ $\log_{L_{ex}} L_{tr}$ & \textbf{21.6} & \textbf{23.0} & \textbf{22.1} & \textbf{22.0} & \textbf{20.6} & \textbf{24.3} \\ \hline
    w/ \emph{truncation} & 20.0 & 19.2 & 19.3 & 19.2 & 17.1 & 21.4 \\
    \hline\hline
    \end{tabular}}
    \quad\quad
    \scalebox{0.9}{
    \begin{tabular}{@{\extracolsep{2pt}}lcccccc}
    \hline\hline
    \multicolumn{7}{c}{\textbf{Code Completion Edit Similarity}}\\
    \hline
    \multirow{2}{*}{Models} & \multicolumn{6}{c}{Sequence Length ($L_{ex}$)} \\
    \cline{2-7}
     & 1k & 2k & 3k & 4k & 5k & 6k \\ \hline
    CodeT5+ & 62.4 & 59.6 & 53.1 & 38.9 & 18.3 & 10.4 \\
    w/ ${\rm P}_{\max}$ & 65.9 & 65.7 & 65.3 & 65.6 & \textbf{63.1} & 64.9 \\
    w/ $\rm H$ & 64.8 & 62.5 & 54.1 & 43.0 & 43.0 & 44.8 \\
    w/ $\log_{L_{ex}} L_{tr}$ & \textbf{66.3} & \textbf{66.1} & \textbf{65.2} & \textbf{66.4} & 63.0 & 66.1 \\ \hline
    w/ \emph{truncation} & 65.3 & 64.2 & 64.2 & 65.6 & 62.2 & \textbf{66.9} \\
    \hline\hline
    \end{tabular}}
    \caption{\textbf{Code Completion Performance.} Full score is 100. We set $L_{tr}=768$. The bucket \emph{n}k contains the data with length in [\emph{n}k, (\emph{n}+1)k), $n\in[1,6]$. For example, the bucket 3k contains data with length in [3000, 4000). See Table~\ref{tab:code_em_full} and~\ref{tab:code_es_full} in Appendix~\ref{sec:appendix_code_exp_full} for the full performance breakdown of $L_{ex}$ up to 16k tokens.}
    \label{tab:code_exp_truncated}
\end{table*}

\subsection{Overall Observations}
First, the maximum probability alignment strategy is the most reliable and best-performing method across most tasks and settings, echoing our discussion in \S\ref{sec:why_retrieval}: For most data, only a subset of the input is useful for a model process at a time. The maximum probability alignment strategy captures this characteristic naturally, thereby outperforming the entropy alignment strategy that cares more about the holistic distribution.

Second, deciding the optimal temperature solely based on sequence lengths, e.g. $\tau=\log_{L_{ex}} L_{tr}$, is not robust enough. 
For example, the perplexity of $\log_{L_{ex}} L_{tr}$ suddenly increases (worse) on T5-XXL-LM, in Table~\ref{tab:lm_exp}, while the other strategies maintain stable results. As another example, it fails to improve the retrieval performance on the Flan-T5-Large model, shown in Table~\ref{tab:retrieval_exp}.

\section{Theoretical Analysis}
\label{sec:analysis}
\subsection{Assumptions}
To shed more light on the underlying mechanisms of the two alignment strategies, we establish the connection between the softmax temperature $\tau$ and data distribution under empirically verified assumptions.
We focus on the 0-th layer (closest to the input embeddings) and take the average over all logit vectors across attention heads.
Note that this is just a crude approximation of Algorithm~\ref{alg:alignment} for analysis purposes since 1) a Transformer language model typically encompasses multiple layers, and 2) in Algorithm~\ref{alg:alignment}, we take the maximum probability or entropy of individual logit vectors as opposed to the average one.

\begin{assumption}
    The length $L$ average logit vector is normally distributed, i.e., its entry $l_i\sim N(0,\sigma^2)$.
    \label{assumption:logit_normal}
\end{assumption}
To compute the~\emph{average logit vector}, we start with a input sequence of length $L$. Using a Transformer model with $H$ attention heads (specifically, a T5 Encoder in our context), we generate $H\times L$ pre-softmax logit vectors, each with a length of $L$. Here, the number of layers is 1 because we focus on the 0-th layer. These logit vectors are then individually sorted, and we subsequently calculate the average of all $H\times L$ sorted logit vectors, resulting in the average logit vector of length $L$.

To assess whether the average logit entries follow a Gaussian distribution, we make use of QQ plots, as illustrated in Figure~\ref{fig:qq_plot}. The linearity of the plot serves as an indicator – the closer the points are to the identity line, the more Gaussian the distribution.
\begin{figure}[!htbp]
    \centering
    \includegraphics[width=0.85\linewidth]{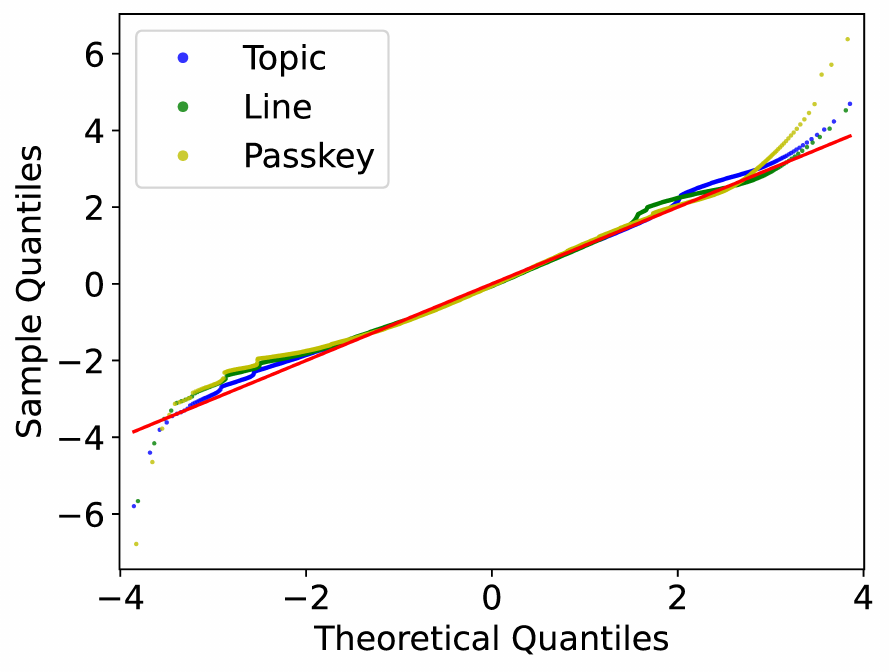}
    \caption{\textbf{QQ plots of Flan-T5-XL.} We experiment with short and long sequences. The red reference line is y=x. We use sequences of length around 512 for this plot. The plot for sequences of length around 15k looks highly similar. Please refer to Appendix~\ref{sec:appendix_qq_plots} for details.}
    \label{fig:qq_plot}
\end{figure}

\vspace{-4mm}
\begin{assumption}
    The largest logit entry of the average logit vector during training and extrapolation is the same: $l_{\max}^{ex}=l_{\max}^{tr}$. See Table~\ref{tab:same_logit}.
    \label{assumption:same_logit}
\end{assumption}
\begin{table}[!htbp]
    \centering\resizebox{\linewidth}{!}{
    \scalebox{0.9}{
    \begin{tabular}{@{\extracolsep{3pt}}ccccccc}
    \hline\hline
    \multicolumn{7}{c}{\textbf{Retrieval Tasks}}\\
    \hline
    \multirow{2}{*}{Criteria} & \multicolumn{2}{c}{Topic} & \multicolumn{2}{c}{Line} & \multicolumn{2}{c}{Passkey} \\
    \cline{2-3} \cline{4-5} \cline{6-7}
     & 512 & 15k & 512 & 15k & 512 & 15k \\ \hline
    $l_{\max}$ & 8.61 & 8.80 & 8.71 & 8.97 & 8.75 & 8.85 \\
    \hline\hline
    \end{tabular}}}
    \caption{\textbf{Largest Logit Entry of Flan-T5-XL.} $l_{\max}$ is the largest logit entry of the average logit vector.}
    \label{tab:same_logit}
\end{table}

\subsection{Maximum Probability Alignment}
\begin{proposition}
Under Assumption~\ref{assumption:logit_normal} and ~\ref{assumption:same_logit}, we can adjust the temperature $\tau$ to align the maximum probability ${\rm P}_{\max}^{tr}={\rm P}_{\max}^{ex}$
\begin{equation}
\begin{split}
    \tau &\approx \frac{\log L_{tr} + \log {\rm P}_{\max}^{tr} + \sigma_{tr}^2/2}{\log L_{ex} + \log {\rm P}_{\max}^{tr} + \sigma_{ex}^2/(2\tau^2)}. \\
    &=\frac{B}{A+\frac{C}{\tau^2}} = \frac{B\tau^2}{A\tau^2 + C}.
\end{split} \nonumber
\end{equation}
Assuming $\tau\neq 0$, we solve the quadratic equation $A\tau^2 - B\tau + C=0$ to get $\tau$. We pick the larger root as our final solution. See proof in Appendix~\ref{sec:appendix_max_derivation}.
\label{prop:align-max}
\end{proposition}

\subsection{Entropy Alignment}
\begin{proposition}
Under Assumption~\ref{assumption:logit_normal},
we can adjust the temperature $\tau$ to align the entropy $H_{tr}=H_{ex}$
\begin{equation*}
\tau \approx \frac{\sigma_{ex}}{\sqrt{\sigma_{tr}^2+2\log \frac{L_{ex}}{L_{tr}}}}
\end{equation*}
\label{prop:align-ent}
See proof in Appendix~\ref{sec:appendix_ent_derivation}.
\end{proposition}

\begin{figure}[!htbp]
    \centering
    \includegraphics[width=0.98\linewidth]{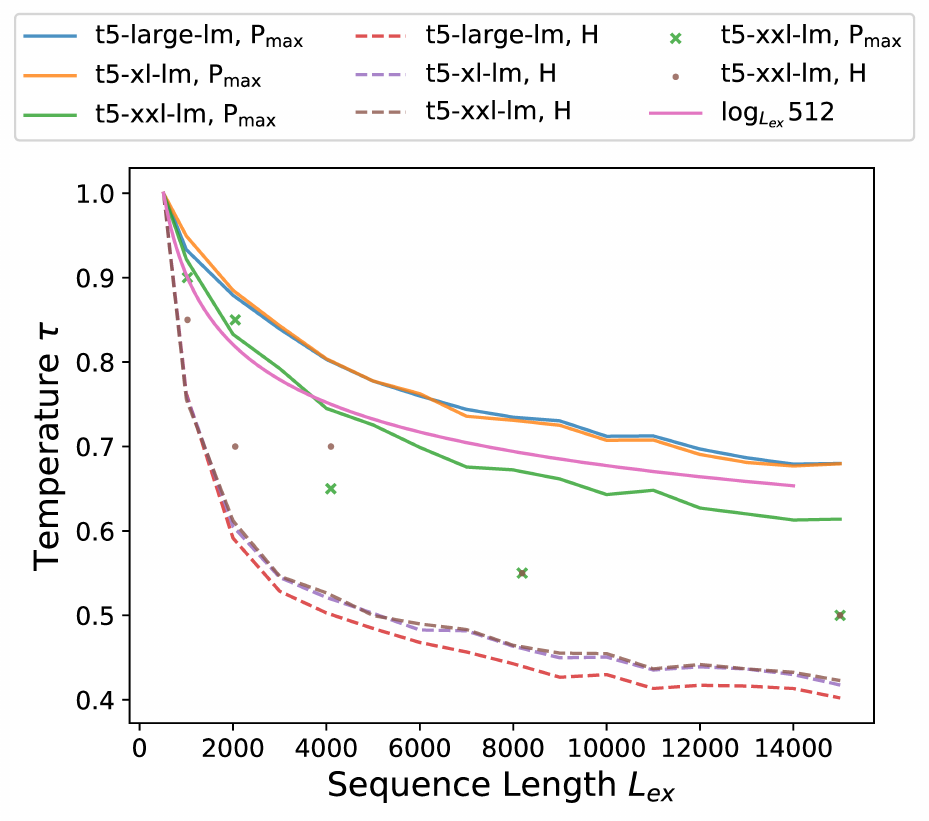}
    \caption{\textbf{Language Modeling Temperature Analysis.} Curves are from Proposition \ref{prop:align-max} \& \ref{prop:align-ent}. Dots and crosses are from Algorithm~\ref{alg:alignment}.}
    \label{fig:lm_temp}
\end{figure}

\begin{figure}[!htbp]
    \centering
    \includegraphics[width=\linewidth]{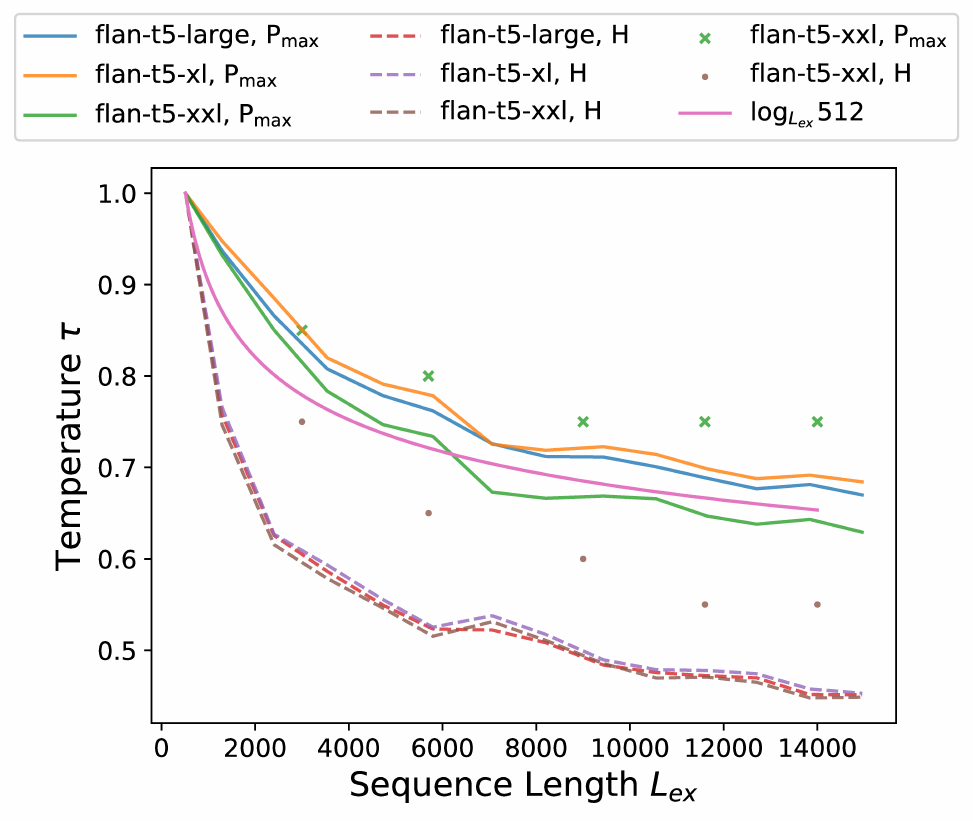}
    \caption{\textbf{Topic Retrieval Temperature Analysis.} Curves are from Proposition \ref{prop:align-max} \& \ref{prop:align-ent}. Dots and crosses are from Algorithm~\ref{alg:alignment}.}
    \label{fig:topic_temp}
\end{figure}

\section{Discussion}
\label{sec:temp_discussion}
The goal of this section is to explain the observations made in \S\ref{sec:experiments} via the lens of temperature analysis.
We visualize Proposition \ref{prop:align-max} and \ref{prop:align-ent} by plotting the temperature curves in Figure~\ref{fig:lm_temp} and ~\ref{fig:topic_temp}. We evaluate ${\rm P}_{\max}^{tr}$ and $\sigma_{tr}$ at the training length and $\sigma_{ex}$ at every extrapolation length considering only the 0-th layer.
You may find the temperature curves for the other tasks in Appendix~\ref{sec:appendix_more_retrieval_plots}.

First, while both proposed strategies lower the temperature when the input sequence length increases, the entropy alignment strategy does so more aggressively, possibly leading to its inferior performance observed in Table~\ref{tab:retrieval_exp} and~\ref{tab:qa_exp} (w/ H). This can be seen by comparing the curves from Propositions \ref{prop:align-max} and \ref{prop:align-ent} or dots from Algorithm~\ref{alg:alignment}.

Second, deciding the optimal temperature based on sequence lengths, e.g. $\tau=\log_{L_{ex}} L_{tr}$, is not the most robust method.
It gives too high of a temperature in Figure~\ref{fig:lm_temp} compared to Algorithm~\ref{alg:alignment}. In other words, it does not sharpen the distribution enough, possibly explaining its perplexity spike in Table~\ref{tab:lm_exp}.
On the other hand, it overly lowers the temperature in Figure~\ref{fig:topic_temp}, thereby failing to improve the retrieval performance on Flan-T5-Large in Table~\ref{tab:retrieval_exp}.

\section{Conclusion}
In this paper, we show that the T5 model family has great potential when it comes to Transformer length extrapolation. We propose the maximum probability and entropy alignment strategies to fix T5's dispersed attention issue without model fine-tuning. We conduct experiments on natural language modeling, retrieval, multi-document question answering, and code completion tasks to demonstrate the effectiveness of our proposed methods. Finally, we present a simplified theoretical analysis to elucidate how the temperature is scaled to achieve attention alignment. We hope that our work can inspire future length-extrapolatable Transformer designs.

\section{Limitations}
\label{sec:limitation}
We base our theoretical analysis on a simplified Transformer language model, which might be further improved by taking all the layers and their interactions into account. 
In addition, we find that different layers have different degrees of distribution flatness, which could be leveraged in future work to perform per-layer fine-grained attention alignment. Finally, our temperature scaling scheme sometimes sharpens a distribution too aggressively in the multi-document question-answering and code completion experiments. This drawback could be possibly improved by designing a more fine-grained attention alignment strategy.

\section*{Acknowledgment}
The first author acknowledges the support from the Boeing Company (2019-STU-PA-259).

\bibliography{custom}

\begin{thebibliography}{34}
\expandafter\ifx\csname natexlab\endcsname\relax\def\natexlab#1{#1}\fi

\bibitem[{AI(2023)}]{XGen}
Salesforce AI. 2023.
\newblock \href {https://blog.salesforceairesearch.com/xgen} {Long sequence modeling with xgen: A 7b llm trained on 8k input sequence length}.

\bibitem[{Biderman et~al.(2023)Biderman, Schoelkopf, Anthony, Bradley, O’Brien, Hallahan, Khan, Purohit, Prashanth, Raff et~al.}]{biderman2023pythia}
Stella Biderman, Hailey Schoelkopf, Quentin~Gregory Anthony, Herbie Bradley, Kyle O’Brien, Eric Hallahan, Mohammad~Aflah Khan, Shivanshu Purohit, USVSN~Sai Prashanth, Edward Raff, et~al. 2023.
\newblock Pythia: A suite for analyzing large language models across training and scaling.
\newblock In \emph{International Conference on Machine Learning}, pages 2397--2430. PMLR.

\bibitem[{Borgeaud et~al.(2022)Borgeaud, Mensch, Hoffmann, Cai, Rutherford, Millican, Van Den~Driessche, Lespiau, Damoc, Clark et~al.}]{retro}
Sebastian Borgeaud, Arthur Mensch, Jordan Hoffmann, Trevor Cai, Eliza Rutherford, Katie Millican, George~Bm Van Den~Driessche, Jean-Baptiste Lespiau, Bogdan Damoc, Aidan Clark, et~al. 2022.
\newblock Improving language models by retrieving from trillions of tokens.
\newblock In \emph{International conference on machine learning}, pages 2206--2240. PMLR.

\bibitem[{Chen et~al.(2023)Chen, Wong, Chen, and Tian}]{chen2023extending}
Shouyuan Chen, Sherman Wong, Liangjian Chen, and Yuandong Tian. 2023.
\newblock Extending context window of large language models via positional interpolation.
\newblock \emph{arXiv preprint arXiv:2306.15595}.

\bibitem[{Chi et~al.(2022)Chi, Fan, Ramadge, and Rudnicky}]{chi2022kerple}
Ta-Chung Chi, Ting-Han Fan, Peter~J Ramadge, and Alexander~I Rudnicky. 2022.
\newblock \href{https://arxiv.org/abs/2205.09921}{KERPLE: Kernelized Relative Positional Embedding for Length Extrapolation}.
\newblock In \emph{Advances in Neural Information Processing Systems (NeurIPS)}, New Orleans, USA.

\bibitem[{Chi et~al.(2023)Chi, Fan, Rudnicky, and Ramadge}]{sandwich}
Ta-Chung Chi, Ting-Han Fan, Alexander~I Rudnicky, and Peter~J Ramadge. 2023.
\newblock \href{https://arxiv.org/abs/2212.10356}{Dissecting Transformer Length Extrapolation via the Lens of Receptive Field Analysis}.
\newblock In \emph{Annual Meeting of the Association for Computational Linguistics (ACL)}, Toronto, Canada.

\bibitem[{Dao et~al.(2022)Dao, Fu, Ermon, Rudra, and R{\'e}}]{dao2022flashattention}
Tri Dao, Dan Fu, Stefano Ermon, Atri Rudra, and Christopher R{\'e}. 2022.
\newblock Flashattention: Fast and memory-efficient exact attention with io-awareness.
\newblock \emph{Advances in Neural Information Processing Systems}, 35:16344--16359.

\bibitem[{Du et~al.(2022)Du, Qian, Liu, Ding, Qiu, Yang, and Tang}]{du2022glm}
Zhengxiao Du, Yujie Qian, Xiao Liu, Ming Ding, Jiezhong Qiu, Zhilin Yang, and Jie Tang. 2022.
\newblock Glm: General language model pretraining with autoregressive blank infilling.
\newblock In \emph{Proceedings of the 60th Annual Meeting of the Association for Computational Linguistics (Volume 1: Long Papers)}, pages 320--335.

\bibitem[{Guo et~al.(2023)Guo, Xu, Duan, Yin, and McAuley}]{Guo2023LongCoderAL}
Daya Guo, Canwen Xu, Nan Duan, Jian Yin, and Julian McAuley. 2023.
\newblock \href {https://api.semanticscholar.org/CorpusID:259262301} {Longcoder: A long-range pre-trained language model for code completion}.
\newblock In \emph{International Conference on Machine Learning}.

\bibitem[{Guu et~al.(2020)Guu, Lee, Tung, Pasupat, and Chang}]{realm}
Kelvin Guu, Kenton Lee, Zora Tung, Panupong Pasupat, and Mingwei Chang. 2020.
\newblock Retrieval augmented language model pre-training.
\newblock In \emph{International conference on machine learning}, pages 3929--3938. PMLR.

\bibitem[{Izacard and Grave(2021)}]{fid}
Gautier Izacard and Edouard Grave. 2021.
\newblock \href {https://doi.org/10.18653/v1/2021.eacl-main.74} {Leveraging passage retrieval with generative models for open domain question answering}.
\newblock In \emph{Proceedings of the 16th Conference of the European Chapter of the Association for Computational Linguistics: Main Volume}, pages 874--880, Online. Association for Computational Linguistics.

\bibitem[{Kwiatkowski et~al.(2019)Kwiatkowski, Palomaki, Redfield, Collins, Parikh, Alberti, Epstein, Polosukhin, Devlin, Lee et~al.}]{kwiatkowski2019natural}
Tom Kwiatkowski, Jennimaria Palomaki, Olivia Redfield, Michael Collins, Ankur Parikh, Chris Alberti, Danielle Epstein, Illia Polosukhin, Jacob Devlin, Kenton Lee, et~al. 2019.
\newblock Natural questions: a benchmark for question answering research.
\newblock \emph{Transactions of the Association for Computational Linguistics}, 7:453--466.

\bibitem[{Lewis et~al.(2020)Lewis, Perez, Piktus, Petroni, Karpukhin, Goyal, K{\"u}ttler, Lewis, Yih, Rockt{\"a}schel et~al.}]{rag}
Patrick Lewis, Ethan Perez, Aleksandra Piktus, Fabio Petroni, Vladimir Karpukhin, Naman Goyal, Heinrich K{\"u}ttler, Mike Lewis, Wen-tau Yih, Tim Rockt{\"a}schel, et~al. 2020.
\newblock Retrieval-augmented generation for knowledge-intensive nlp tasks.
\newblock \emph{Advances in Neural Information Processing Systems}, 33:9459--9474.

\bibitem[{Li et~al.(2023)Li, Shao, Xie, Sheng, Zheng, Gonzalez, Stoica, Ma, and Zhang}]{li2023long}
Dacheng Li, Rulin Shao, Anze Xie, Ying Sheng, Lianmin Zheng, Joseph~E Gonzalez, Ion Stoica, Xuezhe Ma, and Hao Zhang. 2023.
\newblock \href {https://lmsys.org/blog/2023-06-29-longchat/} {How long can opensource llms truly promise on context length}.

\bibitem[{Liu et~al.(2023)Liu, Lin, Hewitt, Paranjape, Bevilacqua, Petroni, and Liang}]{liu2023lost}
Nelson~F Liu, Kevin Lin, John Hewitt, Ashwin Paranjape, Michele Bevilacqua, Fabio Petroni, and Percy Liang. 2023.
\newblock Lost in the middle: How language models use long contexts.
\newblock \emph{arXiv preprint arXiv:2307.03172}.

\bibitem[{Mohtashami and Jaggi(2023)}]{mohtashami2023landmark}
Amirkeivan Mohtashami and Martin Jaggi. 2023.
\newblock Landmark attention: Random-access infinite context length for transformers.
\newblock \emph{arXiv preprint arXiv:2305.16300}.

\bibitem[{Penedo et~al.(2023)Penedo, Malartic, Hesslow, Cojocaru, Cappelli, Alobeidli, Pannier, Almazrouei, and Launay}]{penedo2023refinedweb}
Guilherme Penedo, Quentin Malartic, Daniel Hesslow, Ruxandra Cojocaru, Alessandro Cappelli, Hamza Alobeidli, Baptiste Pannier, Ebtesam Almazrouei, and Julien Launay. 2023.
\newblock The refinedweb dataset for falcon llm: outperforming curated corpora with web data, and web data only.
\newblock \emph{arXiv preprint arXiv:2306.01116}.

\bibitem[{Press et~al.(2022)Press, Smith, and Lewis}]{alibi}
Ofir Press, Noah Smith, and Mike Lewis. 2022.
\newblock \href {https://openreview.net/forum?id=R8sQPpGCv0} {Train short, test long: Attention with linear biases enables input length extrapolation}.
\newblock In \emph{International Conference on Learning Representations}.

\bibitem[{Rabe and Staats(2021)}]{rabe2021self}
Markus~N Rabe and Charles Staats. 2021.
\newblock Self-attention does not need o ($n^2$) memory.
\newblock \emph{arXiv preprint arXiv:2112.05682}.

\bibitem[{Raffel et~al.(2020)Raffel, Shazeer, Roberts, Lee, Narang, Matena, Zhou, Li, and Liu}]{raffel2020exploring}
Colin Raffel, Noam Shazeer, Adam Roberts, Katherine Lee, Sharan Narang, Michael Matena, Yanqi Zhou, Wei Li, and Peter~J Liu. 2020.
\newblock Exploring the limits of transfer learning with a unified text-to-text transformer.
\newblock \emph{The Journal of Machine Learning Research}, 21(1):5485--5551.

\bibitem[{Raychev et~al.(2016)Raychev, Bielik, and Vechev}]{raychev2016probabilistic}
Veselin Raychev, Pavol Bielik, and Martin Vechev. 2016.
\newblock Probabilistic model for code with decision trees.
\newblock \emph{ACM SIGPLAN Notices}, 51(10):731--747.

\bibitem[{Roziere et~al.(2023)Roziere, Gehring, Gloeckle, Sootla, Gat, Tan, Adi, Liu, Remez, Rapin et~al.}]{roziere2023code}
Baptiste Roziere, Jonas Gehring, Fabian Gloeckle, Sten Sootla, Itai Gat, Xiaoqing~Ellen Tan, Yossi Adi, Jingyu Liu, Tal Remez, J{\'e}r{\'e}my Rapin, et~al. 2023.
\newblock Code llama: Open foundation models for code.
\newblock \emph{arXiv preprint arXiv:2308.12950}.

\bibitem[{Scao et~al.(2022)Scao, Fan, Akiki, Pavlick, Ili{\'c}, Hesslow, Castagn{\'e}, Luccioni, Yvon, Gall{\'e} et~al.}]{scao2022bloom}
Teven~Le Scao, Angela Fan, Christopher Akiki, Ellie Pavlick, Suzana Ili{\'c}, Daniel Hesslow, Roman Castagn{\'e}, Alexandra~Sasha Luccioni, Fran{\c{c}}ois Yvon, Matthias Gall{\'e}, et~al. 2022.
\newblock Bloom: A 176b-parameter open-access multilingual language model.
\newblock \emph{arXiv preprint arXiv:2211.05100}.

\bibitem[{Su(2021)}]{kexuefm-8823}
Jianlin Su. 2021.
\newblock \href {https://spaces.ac.cn/archives/8823} {Scaling attention via the lens of entropy invariance}.

\bibitem[{Su et~al.(2021)Su, Lu, Pan, Murtadha, Wen, and Liu}]{rotary}
Jianlin Su, Yu~Lu, Shengfeng Pan, Ahmed Murtadha, Bo~Wen, and Yunfeng Liu. 2021.
\newblock Roformer: Enhanced transformer with rotary position embedding.
\newblock \emph{arXiv preprint arXiv:2104.09864}.

\bibitem[{Svyatkovskiy et~al.(2020)Svyatkovskiy, Deng, Fu, and Sundaresan}]{svyatkovskiy2020intellicode}
Alexey Svyatkovskiy, Shao~Kun Deng, Shengyu Fu, and Neel Sundaresan. 2020.
\newblock Intellicode compose: Code generation using transformer.
\newblock In \emph{Proceedings of the 28th ACM Joint Meeting on European Software Engineering Conference and Symposium on the Foundations of Software Engineering}, pages 1433--1443.

\bibitem[{Team(2023)}]{mpt}
The MosaicML~NLP Team. 2023.
\newblock Introducing mpt-7b: A new standard for open-source, commercially usable llms.
\newblock \emph{https://www.mosaicml.com/blog/mpt-30b}.

\bibitem[{Touvron et~al.(2023)Touvron, Lavril, Izacard, Martinet, Lachaux, Lacroix, Rozi{\`e}re, Goyal, Hambro, Azhar et~al.}]{touvron2023llama}
Hugo Touvron, Thibaut Lavril, Gautier Izacard, Xavier Martinet, Marie-Anne Lachaux, Timoth{\'e}e Lacroix, Baptiste Rozi{\`e}re, Naman Goyal, Eric Hambro, Faisal Azhar, et~al. 2023.
\newblock Llama: Open and efficient foundation language models.
\newblock \emph{arXiv preprint arXiv:2302.13971}.

\bibitem[{Vaswani et~al.(2017)Vaswani, Shazeer, Parmar, Uszkoreit, Jones, Gomez, Kaiser, and Polosukhin}]{vaswani2017attention}
Ashish Vaswani, Noam Shazeer, Niki Parmar, Jakob Uszkoreit, Llion Jones, Aidan~N Gomez, {\L}ukasz Kaiser, and Illia Polosukhin. 2017.
\newblock Attention is all you need.
\newblock \emph{Advances in neural information processing systems}, 30.

\bibitem[{Wang et~al.(2023)Wang, Le, Gotmare, Bui, Li, and Hoi}]{wang2023codet5+}
Yue Wang, Hung Le, Akhilesh~Deepak Gotmare, Nghi~DQ Bui, Junnan Li, and Steven~CH Hoi. 2023.
\newblock Codet5+: Open code large language models for code understanding and generation.
\newblock \emph{arXiv preprint arXiv:2305.07922}.

\bibitem[{Wilk and Gnanadesikan(1968)}]{wilk1968probability}
Martin~B Wilk and Ram Gnanadesikan. 1968.
\newblock Probability plotting methods for the analysis for the analysis of data.
\newblock \emph{Biometrika}, 55(1):1--17.

\bibitem[{Yao et~al.(2021)Yao, Peng, Papadimitriou, and Narasimhan}]{yao-etal-2021-self}
Shunyu Yao, Binghui Peng, Christos Papadimitriou, and Karthik Narasimhan. 2021.
\newblock \href {https://doi.org/10.18653/v1/2021.acl-long.292} {Self-attention networks can process bounded hierarchical languages}.
\newblock In \emph{Proceedings of the 59th Annual Meeting of the Association for Computational Linguistics and the 11th International Joint Conference on Natural Language Processing (Volume 1: Long Papers)}, pages 3770--3785, Online. Association for Computational Linguistics.

\bibitem[{Zeng et~al.(2022)Zeng, Liu, Du, Wang, Lai, Ding, Yang, Xu, Zheng, Xia et~al.}]{zeng2022glm}
Aohan Zeng, Xiao Liu, Zhengxiao Du, Zihan Wang, Hanyu Lai, Ming Ding, Zhuoyi Yang, Yifan Xu, Wendi Zheng, Xiao Xia, et~al. 2022.
\newblock Glm-130b: An open bilingual pre-trained model.
\newblock \emph{arXiv preprint arXiv:2210.02414}.

\bibitem[{Zhang et~al.(2022)Zhang, Roller, Goyal, Artetxe, Chen, Chen, Dewan, Diab, Li, Lin et~al.}]{zhang2022opt}
Susan Zhang, Stephen Roller, Naman Goyal, Mikel Artetxe, Moya Chen, Shuohui Chen, Christopher Dewan, Mona Diab, Xian Li, Xi~Victoria Lin, et~al. 2022.
\newblock Opt: Open pre-trained transformer language models.
\newblock \emph{arXiv preprint arXiv:2205.01068}.

\end{thebibliography}

\appendix
\clearpage

\section{Appendix}
\begin{figure*}[!htbp]
    \centering
    \subfloat[\centering Short sequences around 512]{{\includegraphics[width=0.4\linewidth]{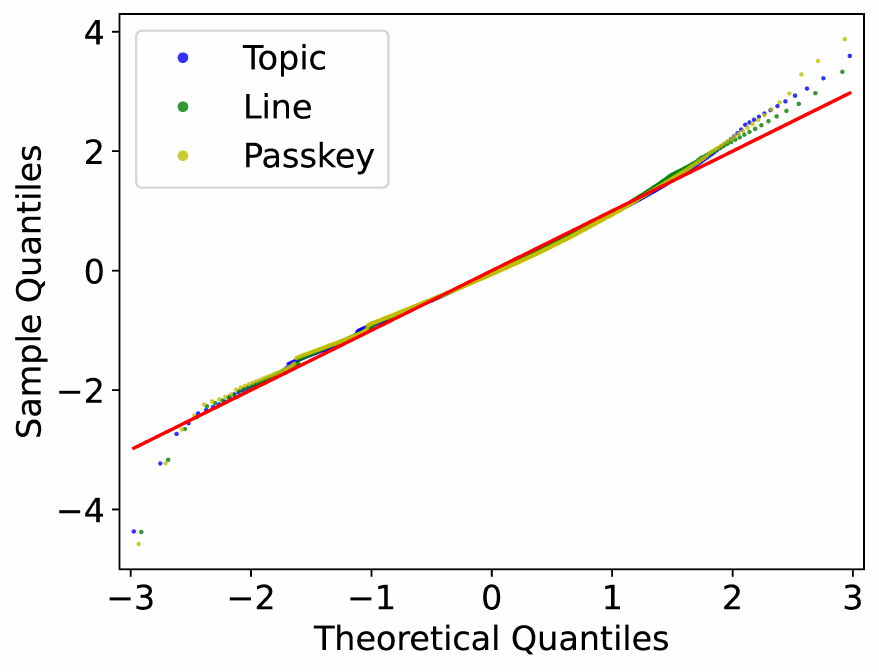}}}
    \quad\quad
    \subfloat[ \centering Long sequences around 15k]{{\includegraphics[width=0.405\linewidth]{qq_long_xl.pdf}}}
    \caption{\textbf{QQ plots of Flan-T5-XL.} We experiment with short and long sequences. The red reference line is y=x. The more closely the scatter plots follow the red reference line, the more Gaussian they are.}
    \label{fig:qq_plots}
\end{figure*}

\subsection{QQ Plots for Assumption~\ref{assumption:logit_normal}}
\label{sec:appendix_qq_plots}
A QQ plot~\cite{wilk1968probability} is a graphical technique used for comparing two probability distributions by plotting their quantiles against each other. A point (x, y) corresponds to a quantile from the second distribution (y-coordinate) plotted against the same quantile from the first distribution (x-coordinate).
When the two distributions under comparison are similar, the points in the QQ plot will roughly align with the identity line, y = x. In our case, where we aim to determine the degree of Gaussian behavior in the average logit vector, the linearity of the plot serves as an indicator – the closer the points are to the identity line, the more Gaussian the distribution.

We present the QQ plots for two lengths, 512 and 15k, on the three retrieval tasks in Figure~\ref{fig:qq_plots}.
They are all close to the red reference line, indicating that their form is highly Gaussian.

\subsection{Detailed Derivation of Proposition~\ref{prop:align-max}}
\label{sec:appendix_max_derivation}
Let $l_{\max}$ be the largest value in the logit vector $l$. Let $\tau$ be the temperature of the Softmax function. The probability of the largest entry is
\begin{align*}
    {\rm P}_{\max} = \frac{e^{l_{\max}/\tau}}{\sum_{i=1}^L e^{l_i/\tau}}.
\end{align*}
Since Softmax is shift-invariant, the logit vector can always be made zero-mean: $\sum_i l_i=0$. Next, according to Assumption~\ref{assumption:logit_normal}, the denominator of Softmax can be approximated as
\begin{equation}
    \sum_{i=1}^L e^{l_i/\tau}\approx L\cdot \mathbb{E}[e^{l_i/\tau}]=L\cdot e^{\sigma^2/(2\tau^2)}
    \label{eq:denominator}
\end{equation}
This implies ${\rm P}_{\max}$ is approximately
\begin{equation*}
    {\rm P}_{\max}\approx \frac{e^{l_{\max}/\tau}}{L e^{\sigma^2/(2\tau^2)}}
\end{equation*}
During the training stage, the temperature $\tau$ is 1
\begin{equation*}
    {\rm P}_{\max}^{tr}\approx \frac{e^{l_{\max}^{tr}}}{L_{tr} e^{\sigma_{tr}^2/2}},
\end{equation*}
which gives an expression of the largest logit entry during the training stage
\begin{equation}
    l_{\max}^{tr}\approx \log \left({\rm P}_{\max}^{tr}L_{tr }e^{\sigma_{tr}^2/2}\right)
    \label{eq:lmax}
\end{equation}
According to Assumption~\ref{assumption:same_logit}, the largest probability during the extrapolation stage can be simplified as
\begin{equation}
\begin{split}
    &{\rm P}_{\max}^{ex} \approx  \frac{e^{l_{\max}^{ex}/\tau}}{L_{ex} e^{\sigma_{ex}^2/(2\tau^2)}} \overset{\text{A.}~\ref{assumption:same_logit}}{=} \frac{e^{l_{\max}^{tr}/\tau}}{L_{ex} e^{\sigma_{ex}^2/(2\tau^2)}}\\ \nonumber
    &\overset{\eqref{eq:lmax}}{\approx} \frac{\left({\rm P}_{\max}^{tr}L_{tr }e^{\sigma_{tr}^2/2}\right)^{1/\tau}}{L_{ex} e^{\sigma_{ex}^2/(2\tau^2)}}
    \label{eq:p-ex}
\end{split}
\end{equation}
Since $\tau$ is a free parameter during extrapolation, we adjust it to carry out the maximum probability alignment strategy. Rearranging the terms gives Proposition~\ref{prop:align-max}.

\subsection{Detailed Derivation of Proposition~\ref{prop:align-ent}}
\label{sec:appendix_ent_derivation}
The entropy of a discrete probability computed by Softmax is
\begin{equation*}
    H = -\sum_i \frac{e^{l_i/\tau}}{D} \log \frac{e^{l_i/\tau}}{D} = \log D -\frac{\sum_i \frac{l_i}{\tau}e^{l_i/\tau}}{D},
\end{equation*}
where $D=\sum_i e^{l_i/\tau}$ is the denominator of Softmax, which can be approximated using Eq.~\eqref{eq:denominator}. On the other hand, we note that $\sum_i l_i e^{l_i}\approx L\mathbb{E}[le^l]$. When $l\sim N(0,\sigma^2)$, $\mathbb{E}[le^l]$ is approximated as
\begin{equation}
\begin{split}
    &\mathbb{E}[le^l]= \int_{-\infty}^\infty \frac{le^l}{\sigma\sqrt{2\pi}}e^{-\frac{l^2}{2\sigma^2}}dl\\
    &=\int_{-\infty}^\infty \frac{l}{\sigma\sqrt{2\pi}}e^{\frac{2\sigma^2 l-l^2}{2\sigma^2}}dl \\
    &=\int_{-\infty}^\infty \frac{l}{\sigma\sqrt{2\pi}}e^{-\frac{(l-\sigma^2)^2-\sigma^4}{2\sigma^2}}dl \\
    &= e^{\sigma^2/2} \int_{-\infty}^\infty \frac{l}{\sigma\sqrt{2\pi}}e^{-\frac{(l-\sigma^2)^2}{2\sigma^2}}dl \\
    &= e^{\sigma^2/2} \sigma^2
    \label{eq:lel-apx}
\end{split}
\end{equation}
Thus, combining Eq.~\eqref{eq:denominator} and \eqref{eq:lel-apx}, the entropy $H$ is approximated as
\begin{align*}
    H &\approx \log L +\frac{\sigma^2}{2\tau^2} - \frac{L e^{\sigma^2/(2\tau^2)} \frac{\sigma^2}{\tau^2}}{L e^{\sigma^2/(2\tau^2)}}\\
    &= \log L - \frac{\sigma^2}{2\tau^2}
\end{align*}
Since $\tau$ is set to 1 during the training stage, we have $H_{tr}\approx \log L_{tr} - \frac{\sigma_{tr}^2}{2}$. During extrapolation, we align the entropy (i.e., $H_{ex}=H_{tr}$) by adjusting $\tau$.
\begin{equation*}
    \log L_{ex} - \frac{\sigma_{ex}^2}{2\tau^2}\approx H_{ex}=H_{tr}\approx \log L_{tr} - \frac{\sigma_{tr}^2}{2}.
\end{equation*}
Since $\tau$ is a free parameter during extrapolation, we adjust it to apply the entropy alignment strategy. Rearranging the terms gives Proposition~\ref{prop:align-ent}.

\subsection{More Real-world Temperature Plots}
\label{sec:appendix_more_retrieval_plots}
We verify Proposition \ref{prop:align-max} and \ref{prop:align-ent} on the remaining tasks by plotting the temperature curves in Figure~\ref{fig:line_temp},~\ref{fig:passkey_temp},~\ref{fig:qa_temp}, and~\ref{fig:code_temp}. We empirically evaluate $\sigma_{tr}$ at the training length and $\sigma_{ex}$ every extrapolation length considering only the 0-th layer.

The real temperatures given by Algorithm~\ref{alg:alignment} are usually higher than those derived from the two propositions. After checking the per-layer attention distributions, we find that the 0-th layer has flatter distributions compared to higher layers. Because the two propositions are derived based on the 0-th layer and a flatter distribution needs a lower temperature to correct, the temperatures given by them tend to be lower than the ones given by Algorithm~\ref{alg:alignment} that takes the average of temperatures across all layers.

\begin{figure}[!htbp]
    \centering
    \includegraphics[width=\linewidth]{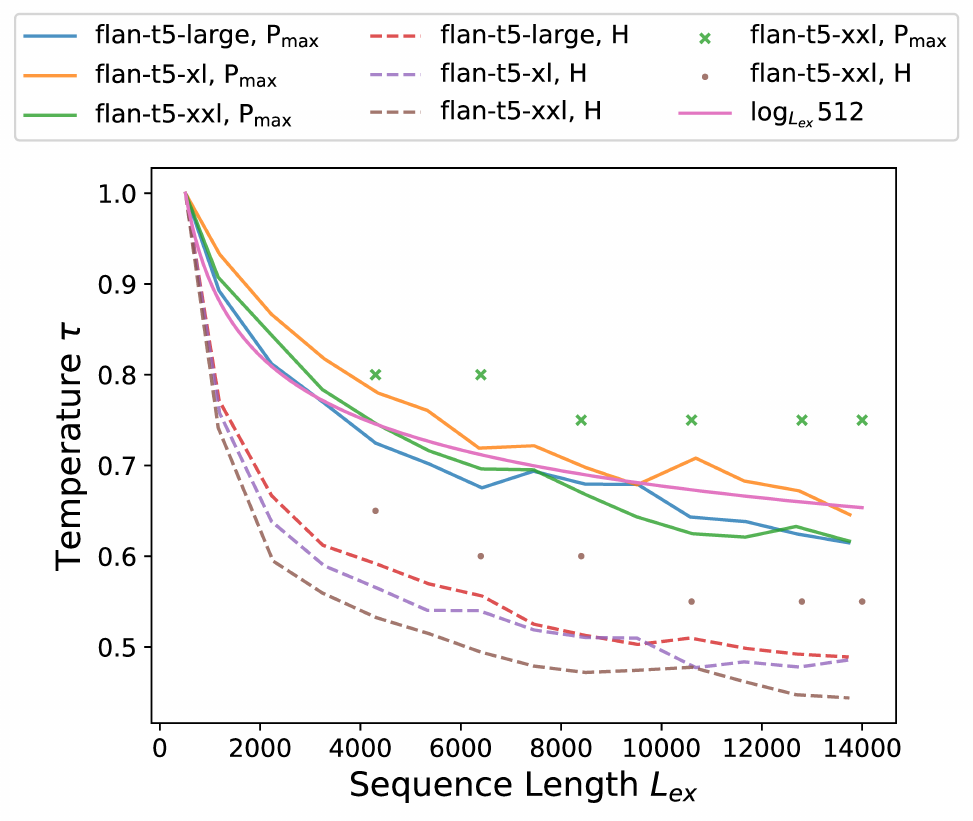}
    \caption{\textbf{Line Retrieval Temperature Analysis.} Curves are given by Proposition~\ref{prop:align-max} and~\ref{prop:align-ent}. Cross signs and dots are given by Algorithm~\ref{alg:alignment}. $\log_L 512$ is given by~\citet{yao-etal-2021-self,kexuefm-8823}.}
    \label{fig:line_temp}
\end{figure}

\begin{figure}[!htbp]
    \centering
    \includegraphics[width=\linewidth]{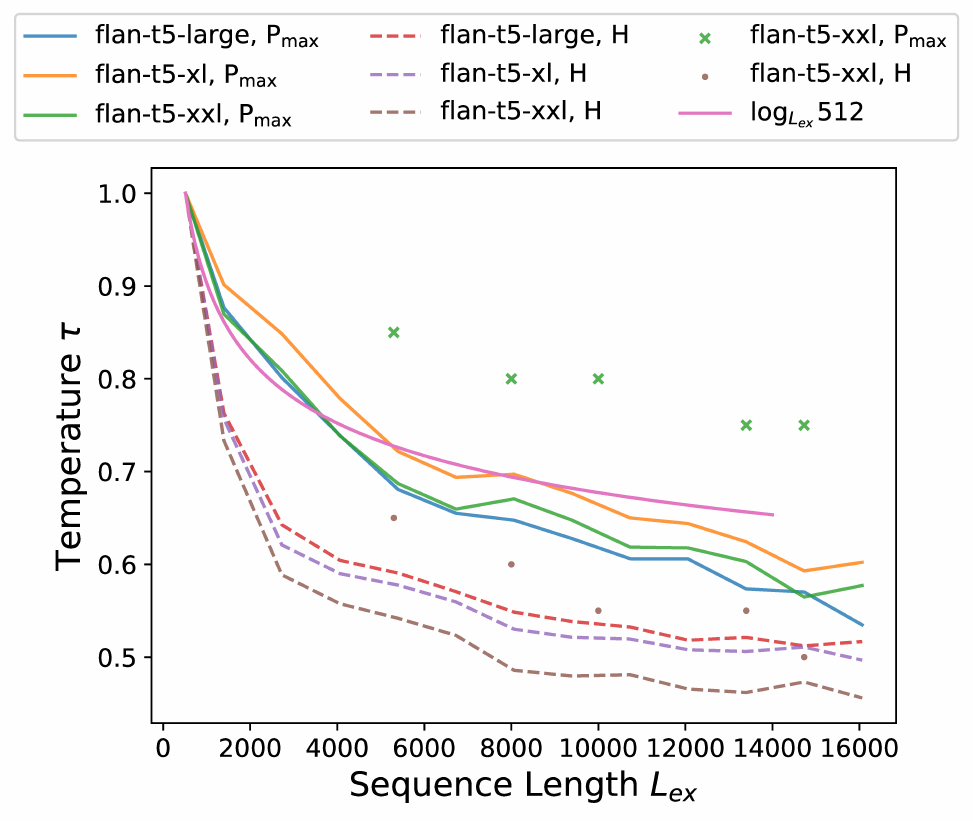}
    \caption{\textbf{Passkey Retrieval Temperature Analysis.} Curves are given by Proposition~\ref{prop:align-max} and~\ref{prop:align-ent}. Cross signs and dots are given by Algorithm~\ref{alg:alignment}. $\log_L 512$ is given by~\citet{yao-etal-2021-self,kexuefm-8823}.}
    \label{fig:passkey_temp}
\end{figure}

\begin{figure}[!htbp]
    \centering
    \includegraphics[width=\linewidth]{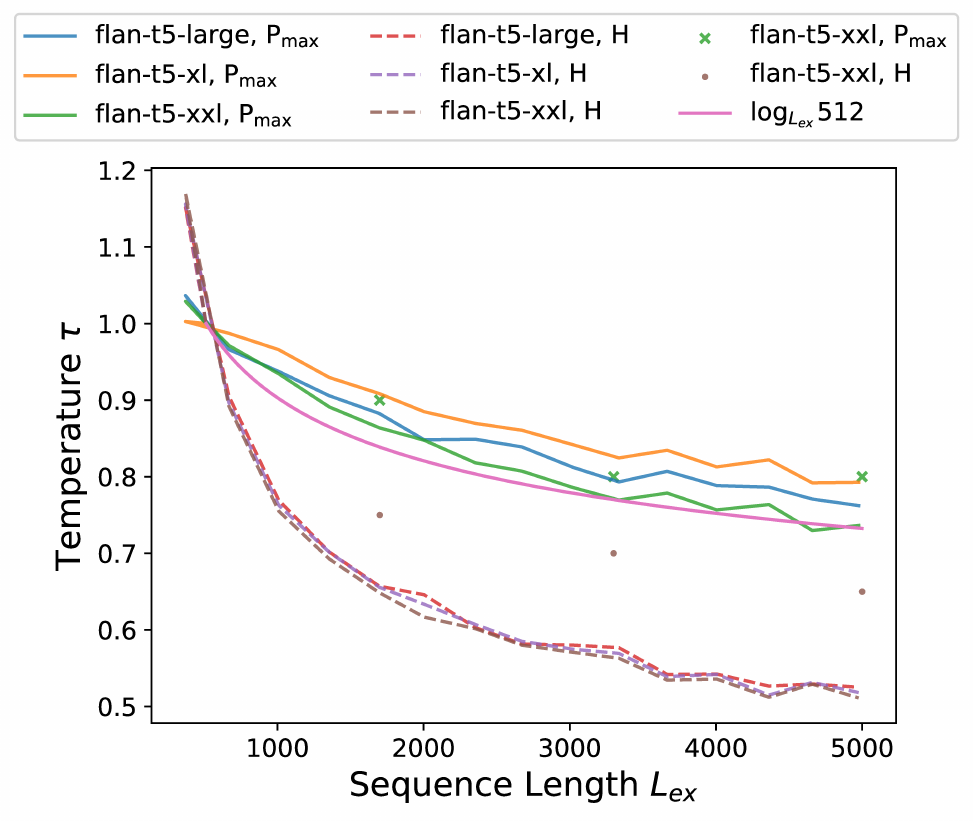}
    \caption{\textbf{Multi-doc QA Temperature Analysis.} Curves are from Proposition \ref{prop:align-max} \& \ref{prop:align-ent}. Dots and crosses are from Algorithm~\ref{alg:alignment}.}
    \label{fig:qa_temp}
\end{figure}

\begin{figure}[!htbp]
    \centering
    \includegraphics[width=0.9\linewidth]{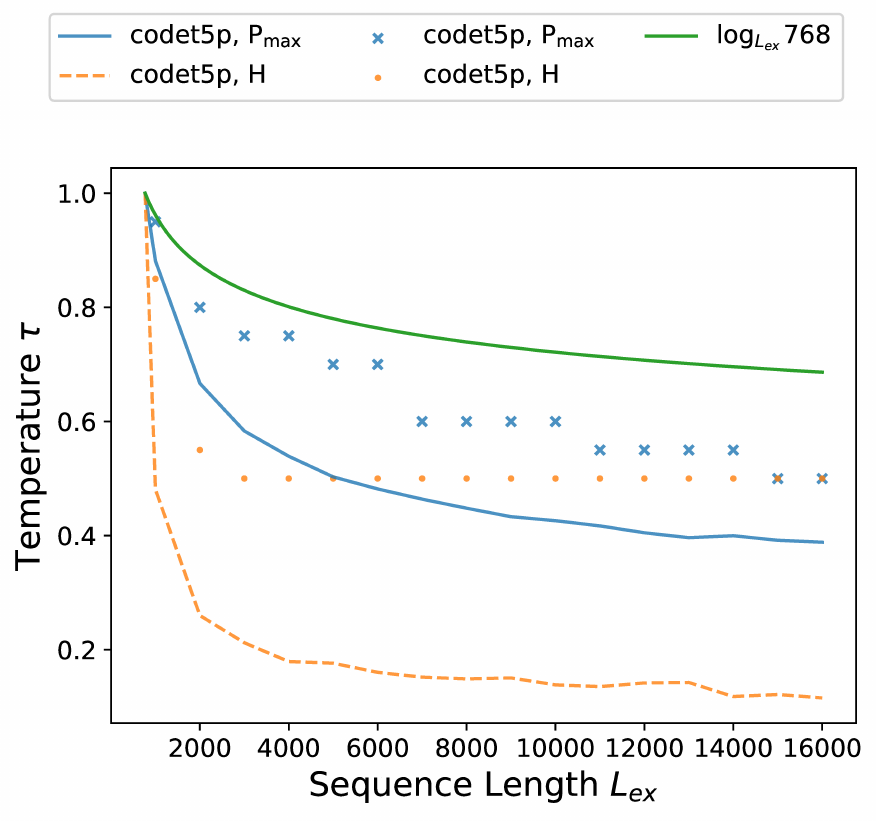}
    \caption{\textbf{Code Completion Temperature Analysis.} Curves are given by Proposition~\ref{prop:align-max} and~\ref{prop:align-ent}. Cross signs and dots are given by Algorithm~\ref{alg:alignment}. $\log_L 768$ is given by~\citet{yao-etal-2021-self,kexuefm-8823}.}
    \label{fig:code_temp}
\end{figure}

\subsection{Detailed Temperature Breakdown}
\label{sec:appendix_temps}
We report the temperatures for all tasks across model sizes given by Algorithm~\ref{alg:alignment} in Table~\ref{tab:lm_temps},~\ref{tab:retrieval_temps},~\ref{tab:qa_temps}, and~\ref{tab:code_temps}.
\begin{table}[!ht]
    \setlength{\tabcolsep}{1.2pt}
    \centering
    \begin{tabular}{@{\extracolsep{1pt}}lccccc}
    \hline\hline
    \multicolumn{6}{c}{\textbf{Language Modeling}}\\
    \hline
    \multirow{2}{*}{Models} & \multicolumn{5}{c}{Sequence Length ($L_{ex}$)} \\
    \cline{2-6}
     & 1024 & 2048 & 4096 & 8192 & 15000 \\ \hline 
    T5-Large-LM \\
    w/ ${\rm P}_{\max}$ & 0.9 & 0.85 & 0.8 & 0.75 & 0.7 \\
    w/ $\rm H$ & 0.8 & 0.7 & 0.6 & 0.5 & 0.5 \\ \hline
    T5-XL-LM \\
    w/ ${\rm P}_{\max}$ & 0.9 & 0.85 & 0.75 & 0.7 & 0.6 \\
    w/ $\rm H$ & 0.85 & 0.7 & 0.55 & 0.5 & 0.5 \\ \hline
    T5-XXL-LM \\
    w/ ${\rm P}_{\max}$ & 0.9 & 0.85 & 0.65 & 0.55 & 0.5 \\
    w/ $\rm H$ & 0.85 & 0.7 & 0.7 & 0.55 & 0.5 \\ \hline
    w/ $\log_{L_{ex}} L_{tr}$ & 0.9 & 0.82 & 0.75 & 0.69 & 0.65 \\
    \hline\hline
    \end{tabular}
    \caption[abc]{\textbf{Temperatures of Language Modeling.} We search the optimal temperature from 1.0, 0.95, 0.9, $\cdots$, 0.5. We set $L_{tr}=512$.}
    \label{tab:lm_temps}
\end{table}

\begin{table*}[!ht]
    \setlength{\tabcolsep}{2pt}
    \centering
    \begin{tabular}{@{\extracolsep{1pt}}lcccccccccccccccc}
    \hline\hline
    \multicolumn{17}{c}{\textbf{Retrieval Tasks}}\\
    \hline
    \multirow{2}{*}{Models} & \multicolumn{5}{c}{Topic, \# of topics} & \multicolumn{6}{c}{Line, \# of lines} & \multicolumn{5}{c}{Passkey, \# of sentences}\\
    \cline{2-6}\cline{7-12}\cline{13-17}
     & 5 & 10 & 15 & 20 & 25 & 200 & 300 & 400 & 500 & 600 & 680 & 20k & 30k & 40k & 50k & 55k \\ \hline
    Flan-T5-Large \\
    w/ ${\rm P}_{\max}$ & 0.85 & 0.8 & 0.75 & 0.75 & 0.75 & 0.85 & 0.8 & 0.8 & 0.75 & 0.75 & 0.75 & 0.85 & 0.80 & 0.80 & 0.75 & 0.75 \\
    w/ $\rm H$ & 0.7 & 0.6 & 0.55 & 0.5 & 0.5 & 0.65 & 0.55 & 0.55 & 0.5 & 0.5 & 0.5 & 0.6 & 0.55 & 0.5 & 0.5 & 0.5 \\ \hline
    Flan-T5-XL \\
    w/ ${\rm P}_{\max}$ & 0.8 & 0.75 & 0.7 & 0.65 & 0.65 & 0.8 & 0.75 & 0.75 & 0.7 & 0.70 & 0.7 & 0.85 & 0.8 & 0.75 & 0.75 & 0.75 \\
    w/ $\rm H$ & 0.7 & 0.55 & 0.55 & 0.5 & 0.5 & 0.6 & 0.55 & 0.55 & 0.5 & 0.5 & 0.5 & 0.7 & 0.65 & 0.6 & 0.6 & 0.6 \\ \hline
    Flan-T5-XXL \\
    w/ ${\rm P}_{\max}$ & 0.85 & 0.8 & 0.75 & 0.75 & 0.75 & 0.8 & 0.8 & 0.75 & 0.75 & 0.75 & 0.75 & 0.85 & 0.8 & 0.8 & 0.75 & 0.75 \\
    w/ $\rm H$ & 0.75 & 0.65 & 0.6 & 0.55 & 0.55 & 0.65 & 0.6 & 0.6 & 0.55 & 0.55 & 0.55 & 0.65 & 0.6 & 0.55 & 0.55 & 0.5 \\ \hline
    w/ $\log_{L_{ex}} L_{tr}$ & 0.79 & 0.72 & 0.69 & 0.67 & 0.65 & 0.74 & 0.71 & 0.69 & 0.67 & 0.66 & 0.65 & 0.73 & 0.69 & 0.67 & 0.66 & 0.65 \\
    \hline\hline
    \end{tabular}
    \caption{\textbf{Temperatures of Retrieval Tasks.} We search the optimal temperature from 1.0, 0.95, 0.9, $\cdots$, 0.5. The maximum lengths of the three tasks are all around 14.5k to 15.5k tokens ($L_{ex}$). We set $L_{tr}=512$.}
    \label{tab:retrieval_temps}
\end{table*}

\begin{table}[!ht]
    \setlength{\tabcolsep}{2pt}
    \centering
    \begin{tabular}{@{\extracolsep{1.2pt}}lcccc}
    \hline\hline
    \multicolumn{4}{c}{\textbf{Multi-document Question Answering}}\\
    \hline
    \multirow{2}{*}{Models} & 10 Docs & 20 Docs & 30 Docs \\
    & {\small $L_{ex}=1700$} & {\small $L_{ex}=3300$} & {\small $L_{ex}=5000$} \\
    \hline
    Flan-T5-Large \\
    w/ Max. & 0.9 & 0.85 & 0.8 \\
    w/ Ent. & 0.75 & 0.65 & 0.6 \\ \hline
    Flan-T5-XL \\
    w/ Max. & 0.85 & 0.75 & 0.75 \\
    w/ Ent. & 0.75 & 0.65 & 0.55 \\ \hline
    Flan-T5-XXL \\
    w/ Max. & 0.9 & 0.8 & 0.8 \\
    w/ Ent. & 0.75 & 0.7 & 0.65 \\ \hline
    w/ $\log_{L_{ex}} L_{tr}$ & 0.84 & 0.77 & 0.73 \\
    \hline\hline
    \end{tabular}
    \caption{\textbf{Temperatures of Multi-document Question Answering.} We search the optimal temperature from 1.0, 0.95, 0.9, $\cdots$, 0.5. Different golden document positions have the same temperature. We set $L_{tr}=512$.}
    \label{tab:qa_temps}
\end{table}

\begin{table*}[!ht]
    \setlength{\tabcolsep}{2pt}
    \centering
    \begin{tabular}{@{\extracolsep{1pt}}lcccccccccccccccc}
    \hline\hline
    \multicolumn{17}{c}{\textbf{Code Completion}}\\
    \hline
    \multirow{2}{*}{Models} & \multicolumn{16}{c}{Sequence Length ($L_{ex}$)} \\
    \cline{2-17}
     & 1k & 2k & 3k & 4k & 5k & 6k & 7k & 8k & 9k & 10k & 11k & 12k & 13k & 14k & 15k & 16k \\ \hline
    CodeT5+ \\
    w/ ${\rm P}_{\max}$ & 0.95 & 0.8 & 0.75 & 0.75 & 0.7 & 0.7 & 0.6 & 0.6 & 0.6 & 0.6 & 0.55 & 0.55 & 0.55 & 0.55 & 0.5 & 0.5 \\
    w/ $\rm H$ & 0.85 & 0.55 & 0.5 & 0.5 & 0.5 & 0.5 & 0.5 & 0.5 & 0.5 & 0.5 & 0.5 & 0.5 & 0.5 & 0.5 & 0.5 & 0.5 \\
    w/ $\log_{L_{ex}} L_{tr}$ & 0.96 & 0.87 & 0.83 & 0.8 & 0.78 & 0.76 & 0.75 & 0.74 & 0.73 & 0.72 & 0.71 & 0.71 & 0.7 & 0.7 & 0.69 & 0.69 \\
    \hline\hline
    \end{tabular}
    \caption{\textbf{Temperatures of Code Completion.} We search the optimal temperature from 1.0, 0.95, 0.9, $\cdots$, 0.5. The maximum length is around 16k tokens ($L_{ex}$). We set $L_{tr}=768$.}
    \label{tab:code_temps}
\end{table*}

\subsection{Performance Breakdown of Code Completion}
\label{sec:appendix_code_exp_full}
We report the performance breakdown of Exact Match and Edit Similarity across lengths in Table~\ref{tab:code_em_full} and~\ref{tab:code_es_full}.

\begin{table*}[!ht]
    \setlength{\tabcolsep}{2pt}
    \centering
    \begin{tabular}{@{\extracolsep{1pt}}lccccccccccccccc}
    \hline\hline
    \multicolumn{16}{c}{\textbf{Code Completion Exact Match}}\\
    \hline
    \multirow{2}{*}{Models} & \multicolumn{15}{c}{Sequence Length ($L_{ex}$)} \\
    \cline{2-16}
     & 1k & 2k & 3k & 4k & 5k & 6k & 7k & 8k & 9k & 10k & 11k & 12k & 13k & 14k & 15k \\ \hline
    CodeT5+ & 19.6 & 19.0 & 11.3 & 2.6 & 0.1 & 0.0 & 0.0 & 0.0 & 0.0 & 0.0 & 0.0 & 0.0 & 0.0 & 0.0 & 0.0 \\
    w/ ${\rm P}_{\max}$ & \textcolor{red}{21.1} & \textcolor{red}{22.5} & \textcolor{red}{21.7} & \textcolor{red}{21.5} & \textcolor{red}{19.3} & \textcolor{red}{22.7} & 16.1 & 14.4 & 13.4 & 20.6 & 16.0 & 15.3 & 12.3 & 16.7 & 4.5 \\
    w/ $\rm H$ & 19.5 & 18.7 & 13.7 & 9.0 & 7.9 & 9.0 & 10.3 & 8.8 & 10.8 & 12.1 & 11.7 & 10.2 & 9.2 & 11.1 & 2.3 \\
    w/ $\log_{L_{ex}} L_{tr}$ & \textcolor{red}{21.6} & \textcolor{red}{23.0} & \textcolor{red}{22.1} & \textcolor{red}{22.0} & \textcolor{red}{20.6} & \textcolor{red}{24.3} & 20.7 & \textcolor{red}{18.6} & 19.1 & 22.4 & 13.8 & 20.3 & 15.4 & 19.4 & 11.4 \\ \hline
    w/ \emph{truncation} & 20.0 & 19.2 & 19.3 & 19.2 & 17.1 & 21.4 & 21.1 & 18.0 & 19.1 & 25.2 & 18.1 & 20.3 & 16.9 & 27.8 & 15.9 \\
    \hline\hline
    \end{tabular}
    \caption{\textbf{Full Exact Match Breakdown of Code Completion Edit Similarity.} We set $L_{tr}=768$. \textcolor{red}{Numbers} in red are higher than their counterpart in the w/\emph{truncation} row. The bucket \emph{n}k contains the data with length in [\emph{n}k, (\emph{n}+1)k), $n\in[1,15]$.}
    \label{tab:code_em_full}
\end{table*}

\begin{table*}[!ht]
    \setlength{\tabcolsep}{2pt}
    \centering
    \begin{tabular}{@{\extracolsep{1pt}}lccccccccccccccc}
    \hline\hline
    \multicolumn{16}{c}{\textbf{Code Completion}}\\
    \hline
    \multirow{2}{*}{Models} & \multicolumn{15}{c}{Sequence Length ($L_{ex}$)} \\
    \cline{2-16}
     & 1k & 2k & 3k & 4k & 5k & 6k & 7k & 8k & 9k & 10k & 11k & 12k & 13k & 14k & 15k \\ \hline
    CodeT5+ & 62.4 & 59.6 & 53.1 & 38.9 & 18.3 & 10.4 & 6.1 & 4.0 & 4.5 & 5.0 & 6.7 & 5.1 & 6.4 & 4.4 & 3.5 \\
    w/ ${\rm P}_{\max}$ & \textcolor{red}{65.9} & \textcolor{red}{65.7} & \textcolor{red}{65.3} & 65.6 & \textcolor{red}{63.1} & 64.9 & 60.0 & 60.0 & 58.1 & 57.5 & 56.2 & 56.0 & 52.1 & 56.9 & 39.9 \\
    w/ $\rm H$ & 64.8 & 62.5 & 54.1 & 43.0 & 43.0 & 44.8 & 47.7 & 47.0 & 47.6 & 51.2 & 44.3 & 49.7 & 50.3 & 57.4 & 42.0 \\
    w/ $\log_{L_{ex}} L_{tr}$ & \textcolor{red}{66.3} & \textcolor{red}{66.1} & \textcolor{red}{65.2} & \textcolor{red}{66.4} & \textcolor{red}{63.0} & 66.1 & 61.9 & 58.8 & 61.6 & 57.8 & 54.2 & 57.9 & 48.7 & 52.2 & 48.6 \\ \hline
    w/ \emph{truncation} & 65.3 & 64.2 & 64.2 & 65.6 & 62.2 & 66.9 & 66.8 & 61.8 & 64.1 & 65.1 & 63.5 & 63.9 & 61.5 & 67.6 & 60.8 \\
    \hline\hline
    \end{tabular}
    \caption{\textbf{Full Edit Similarity Breakdown of Code Completion.} We set $L_{tr}=768$. \textcolor{red}{Numbers} in red are higher than their counterpart in the w/\emph{truncation} row. The bucket \emph{n}k contains the data with length in [\emph{n}k, (\emph{n}+1)k), $n\in[1,15]$.}
    \label{tab:code_es_full}
\end{table*}

\subsection{Performance Breakdown of Multi-document Question Answering}
\label{sec:appendix_qa_exp_full}
We report the performance breakdown of different numbers of input documents in Table~\ref{tab:qa_exp_full}.

\begin{table*}[!ht]
    \setlength{\tabcolsep}{2.5pt}
    \centering
    \begin{tabular}{@{\extracolsep{1.2pt}}lccccccccccccccc}
    \hline\hline
    \multicolumn{16}{c}{\textbf{Multi-document Question Answering}}\\
    \hline
    \multirow{2}{*}{Models} & \multicolumn{3}{c}{10 Docs} & \multicolumn{5}{c}{20 Docs} & \multicolumn{7}{c}{30 Docs} \\
    \cline{2-4}\cline{5-9}\cline{10-16}
     & 0 & 4 & 9 & 0 & 4 & 9 & 14 & 19 & 0 & 4 & 9 & 14 & 19 & 24 & 29 \\ \hline
    Flan-T5-Large & 60.6 & 48.5 & 48.0 & 54.5 & 44.0 & 39.6 & 38.0 & 40.2 & 52.6 & 42.0 & 36.5 & 34.0 & 33.9 & 33.9 & 37.9 \\
    w/ Max. & 60.9 & 49.8 & 48.6 & 53.5 & 45.6 & 40.8 & 39.7 & 41.3 & 50.8 & 44.5 & 39.5 & 36.4 & 35.9 & 35.8 & 37.0 \\
    w/ Ent. & 58.9 & 50.1 & 47.3 & 52.4 & 45.2 & 40.4 & 38.0 & 40.0 & 47.6 & 41.1 & 35.2 & 33.5 & 32.2 & 33.3 & 34.2 \\
    w/ $\log_{L_{ex}} L_{tr}$ & 60.2 & 51.1 & 48.4 & 53.8 & 46.0 & 41.4 & 39.4 & 41.7 & 50.6 & 44.1 & 39.3 & 36.3 & 35.8 & 35.8 & 37.2 \\ \hline
    Flan-T5-XL & 64.0 & 55.4 & 58.9 & 60.6 & 47.9 & 45.1 & 47.3 & 55.3 & 58.4 & 44.6 & 40.0 & 39.9 & 41.7 & 46.4 & 54.8 \\
    w/ Max. & 65.3 & 57.3 & 60.8 & 62.2 & 51.6 & 49.0 & 49.4 & 56.0 & 60.9 & 49.1 & 46.0 & 44.9 & 46.3 & 49.1 & 55.7 \\
    w/ Ent. & 64.7 & 56.7 & 60.0 & 59.3 & 50.1 & 47.9 & 49.8 & 55.1 & 52.4 & 43.5 & 42.1 & 40.3 & 42.0 & 42.9 & 51.3 \\
    w/ $\log_{L_{ex}} L_{tr}$ & 65.1 & 57.0 & 60.6 & 62.2 & 51.7 & 48.8 & 49.5 & 56.0 & 61.0 & 49.1 & 46.1 & 44.7 & 46.1 & 48.7 & 55.4 \\ \hline
    Flan-T5-XXL & 65.1 & 61.0 & 64.6 & 61.1 & 53.9 & 52.4 & 54.7 & 62.4 & 58.9 & 49.1 & 48.1 & 47.5 & 48.9 & 53.1 & 61.2 \\
    w/ Max. & 66.2 & 61.8 & 63.2 & 62.8 & 55.9 & 54.4 & 55.6 & 59.6 & 60.4 & 52.5 & 51.0 & 50.2 & 51.3 & 53.5 & 59.1 \\
    w/ Ent. & 67.3 & 62.1 & 61.3 & 63.2 & 56.1 & 54.1 & 54.3 & 57.6 & 61.0 & 53.4 & 50.8 & 50.3 & 50.7 & 51.9 & 55.7 \\
    w/ $\log_{L_{ex}} L_{tr}$ & 66.7 & 61.9 & 63.1 & 63.1 & 56.0 & 54.7 & 55.1 & 59.0 & 61.5 & 53.3 & 51.3 & 50.3 & 51.1 & 53.0 & 57.2 \\
    \hline\hline
    \end{tabular}
    \caption{\textbf{Full Performance Breakdown of Multi-document Question Answering.} The numbers are accuracy. Full score is 100. 0, 4, 9... indicate the position of the golden document that contains the answer to a question.}
    \label{tab:qa_exp_full}
\end{table*}

\end{document}